\documentclass[10pt,twocolumn,letterpaper]{article}

\usepackage[utf8]{inputenc}
\usepackage{cvpr}
\usepackage{times}
\usepackage{epsfig}
\usepackage{subfig}
\usepackage{graphicx}\graphicspath{{figs/}}
\usepackage{amsmath,amssymb,amsthm,amsfonts}
\usepackage{color}
\usepackage{caption}
\usepackage{xspace}
\usepackage{array}
\usepackage{cite}
\usepackage{ifthen}
\usepackage[pagebackref=true,breaklinks=true,colorlinks,bookmarks=false]{hyperref}
\usepackage[dvipsnames,svgnames,x11names]{xcolor}
\usepackage{booktabs}
\usepackage{multirow}
\usepackage{enumitem}
\usepackage{tabularx}

\newcommand{\bI}{\mathbf{I}}
\newcommand{\bJ}{\mathbf{J}}
\newcommand{\bK}{\mathbf{K}}

\newcommand{\bm}{\mathbf{m}}

\newcommand{\bp}{\mathbf{p}}

\newcommand{\br}{\mathbf{r}}

\newcommand{\bt}{\mathbf{t}}\newcommand{\bT}{\mathbf{T}}

\newcommand{\bW}{\mathbf{W}}
\newcommand{\bx}{\mathbf{x}}

\newcommand{\bzero}{\mathbf{0}}

\newcommand{\blambda}{\boldsymbol{\lambda}}

\newcommand{\bxi}{\boldsymbol{\xi}}

\newcommand{\nL}{\mathbb{L}}

\newcommand{\nR}{\mathbb{R}}

\newcommand{\cL}{\mathcal{L}}

\newcommand{\cP}{\mathcal{P}}

\newcommand{\cW}{\mathcal{W}}

\newcommand{\figref}[1]{Fig.~\ref{#1}}
\newcommand{\secref}[1]{Section~\ref{#1}}

\newcommand{\tabref}[1]{Table~\ref{#1}}

\makeatletter
\DeclareRobustCommand\onedot{\futurelet\@let@token\@onedot}
\def\@onedot{\ifx\@let@token.\else.\null\fi\xspace}
\def\eg{e.g\onedot} 
\def\ie{i.e\onedot}

\def\etal{et~al\onedot}

\makeatother

\newcommand{\boldparagraph}[1]{\vspace{0.2cm}\noindent{\bf #1:} }

\definecolor{darkgreen}{rgb}{0,0.7,0}

\usepackage[breaklinks=true,bookmarks=false]{hyperref}

\cvprfinalcopy 

\def\cvprPaperID{3963} 

\begin{document}

\title{Taking a Deeper Look at the Inverse Compositional Algorithm}

\author{Zhaoyang Lv$^{1,2}$ \quad Frank Dellaert$^{1}$ \quad James M. Rehg$^{1}$ \quad Andreas Geiger$^{2}$\\
	$^{1}$Georgia Institute of Technology, Atlanta, United States\\
	$^{2}$Autonomous Vision Group, MPI-IS and University of Tübingen, Germany\\
	{\tt\small \{zhaoyang.lv, rehg\}@gatech.edu \quad frank.dellaert@cc.gatech.edu \quad andreas.geiger@tue.mpg.de}
}


\maketitle

\begin{abstract}
In this paper, we provide a modern synthesis of the classic inverse compositional algorithm for dense image alignment. We first discuss the assumptions made by this well-established technique, and subsequently propose to relax these assumptions by incorporating data-driven priors into this model. More specifically, we unroll a robust version of the inverse compositional algorithm and replace multiple components of this algorithm using more expressive models whose parameters we train in an end-to-end fashion from data. Our experiments on several challenging 3D rigid motion estimation tasks demonstrate the advantages of combining optimization with learning-based techniques, outperforming the classic inverse compositional algorithm as well as data-driven image-to-pose regression approaches.
\end{abstract}

\newcommand{\loss}{\mathcal{L}}


\newcommand{\rgbd}{\mbox{RGB-D} }

\newcommand{\energy}{\mathbf{E}}
\newcommand{\regularization}{\mathbf{R}}
\newcommand{\values}{\mathbf{V}}
\newcommand{\robustnorm}{\rho}
\newcommand{\weight}{\mathbf{w}}
\newcommand{\weightmatrix}{\mathbf{W}}
\newcommand{\type}{\mathbf{Z}}
\newcommand{\descriptor}[2]{\mathbf{C}_{#1}(#2)}
\newcommand{\residual}{\mathbf{r}}
\newcommand{\jacobian}{\mathbf{J}}
\newcommand{\hessian}{\mathbf{H}}
\newcommand{\damping}{\mathbf{D}}
\newcommand{\diag}[1]{\text{diag}{(#1)}}
\newcommand{\eye}{\mathbf{I}}

\newcommand{\mapto}{\rightarrow}
\newcommand{\composite}{\circ}
\newcommand{\Hadamardproduct}{\circ}
\newcommand{\gradient}{\bigtriangledown}


\newcommand{\image}{I}
\newcommand{\depth}{D}
\newcommand{\flow}{F}
\newcommand{\feature}{\mathcal{F}}
\newcommand{\warp}{\mathcal{W}}

\newcommand{\domain}{\Omega}

\newcommand{\planes}{\mathcal{P}}
\newcommand{\plane}{\bar{\mathbf{n}}}
\newcommand{\motion}{\mathcal{X}}
\newcommand{\edge}{\mathcal{E}}

\newcommand{\point}{\mathbf{x}}

\newcommand{\pixel}{\mathbf{u}}

\newcommand{\vertexmap}{V}

\newcommand{\segment}{S}

\newcommand{\occlusion}{O}

\newcommand{\opticalflow}[2]{\delta \pixel_{{#1}\rightarrow{#2}}}
\newcommand{\sceneflow}[2]{\delta \point_{{#1} \rightarrow {#2}}}

\newcommand{\OF}{\delta \pixel}
\newcommand{\SF}{\delta \point}

\newcommand{\extrinsics}{\mathcal{C}}
\newcommand{\intrinsic}{\mathbf{K}}
\newcommand{\extrinsic}[1]{\mathcal{C}_{#1}}

\newcommand{\transform}[2]{\mathcal{T}_{#1 \rightarrow #2}}
\newcommand{\eulerangle}{\Theta}
\newcommand{\unitaxis}{\omega}

\newcommand{\rotation}{\mathbf{R}}
\newcommand{\translation}{\mathbf{t}}
\newcommand{\project}[1]{\pi(#1)}
\newcommand{\invproject}[1]{\pi^{-1}(#1)}

\newcommand{\homography}{\mathbf{H}}

\newcommand{\R}[1]{\mathbf{R}{(#1)}}
\newcommand{\SO}[1]{\mathbf{SO}{(#1})}
\newcommand{\SE}[1]{\mathbf{SE}{(#1)}}
\newcommand{\se}[1]{\mathfrak{se}{(#1)}}
\newcommand{\so}[1]{\mathfrak{so}{(#1)}}
\newcommand{\M}[1]{\mathbb{R}{^{#1}}}
\newcommand{\expM}[1]{\exp^{[#1]_{\times}}}


\newcommand{\zero}{\mathbf{0}}
\newcommand{\identity}{\mathbf{I}}
\newcommand{\pesudoinverse}{\dagger}

\newcommand{\normtwo}[1]{||#1||_2}
\newcommand{\mask}[1]{M}
\newcommand{\iversonbracket}[1]{[#1]}

\newcommand{\varVec}[1]{\mathbf{#1}}		
\newcommand{\varVecGreek}[1]{\bm{#1}}	
\newcommand{\tp}{^\top}					
\newcommand{\skewsymmetric}[1]{[#1]_{\times}}


\renewcommand{\vec}[1]{\bm{#1}}
\newcommand{\mat}[1]{\bm{#1}}
\newcommand{\img}[1]{\mathcal{#1}}
\newcommand{\transpose}[1]{#1\tp}
\newcommand{\norm}[1]{\left\|#1\right\|}
\newcommand{\inv}[1]{#1^{\hbox{-}1}}

\newcommand{\invProject}{\inv{\pi}}
\newcommand{\rsProject}{\tau}
\newcommand{\voxel}{\vec{v}}  

\newcommand{\refFig}[1]{Figure~\ref{#1}}
\newcommand{\refSubFig}[1]{(\subref{#1})}
\newcommand{\refTab}[1]{Table~\ref{#1}}
\newcommand{\refEq}[1]{(\ref{#1})}
\newcommand{\refSec}[1]{Section~\ref{#1}}
\newcommand{\refList}[1]{Listing~\ref{#1}}

\newcommand{\volume}{\mathop{\ooalign{\hfil$V$\hfil\cr\kern0.08em--\hfil\cr}}\nolimits}
\newcommand{\Vol}{\rotatebox[origin=c]{180}{\ensuremath{A}}}

\section{Introduction}
\label{sec:introduction}

Since the seminal work by Lucas and Kanade \cite{Lucas1981IJCAI}, dense image alignment has become an ubiquitous tool in computer vision with many applications including stereo reconstruction \cite{Scharstein2002IJCV}, tracking \cite{Wang18icra,Bibi2016ECCV,Shi1994CVPR},
image registration \cite{Brown2007IJCV,Levin2004ECCV,Shum2000IJCV}, super-resolution \cite{Irani1991CVGIP} and SLAM \cite{Clark18eccv,Engel14eccv,Engel18pami}.
In this paper, we provide a learning-based perspective on the Inverse Compositional algorithm, an efficient variant of the original Lucas-Kanade image registration technique. In particular, we lift some of the restrictive assumptions by parameterizing several components of the algorithm using neural networks and training the entire optimization process end-to-end.
In order to put contributions into context, we will now briefly review the Lucas-Kanade algorithm, the Inverse Compositional algorithm, as well as the robust M-Estimator which form the basis for our model.
More details can be found in the comprehensive reviews of Baker \etal \cite{Baker2002,Baker2003}.

\boldparagraph{Lucas-Kanade Algorithm}
The Lucas-Kanade algorithm minimizes the photometric error between a template and an image. Letting $\bT:\Xi\rightarrow\nR^{W\times H}$ and $\bI:\Xi\rightarrow\nR^{W\times H}$ denote the warped template and image, respectively%
\footnote{The warping function $\cW_{\bxi}:\nR^2\rightarrow\nR^2$ might represent translation, affine 2D motion or (if depth is available) rigid or non-rigid 3D motion. To avoid clutter in the notation, we do not make $\cW_{\bxi}$ explicit in our equations.}, the Lucas-Kanade objective can be stated as follows
%
%
\begin{equation}
\min_{\bxi} {\Vert \bI(\bxi) - \bT (\bzero)\Vert}_2^2
\label{eq:lk_objective}
\end{equation}
%
where $\bI(\bxi)$ denotes image $\bI$ transformed using warp parameters $\bxi$ and $\bT(\bzero)=\bT$ denotes the original template.

Minimizing \eqref{eq:lk_objective} is a non-linear optimization task as the image $\bI$ depends non-linearly on the warp parameters $\bxi$.
The Lucas-Kanade algorithm therefore iteratively solves for the warp parameters $\bxi_{k+1} = \bxi_{k} \circ \Delta \bxi$. At every iteration $k$, the warp increment $\Delta \bxi$ is obtained by linearizing
%
%
\begin{equation}
\min_{\Delta\bxi} {\Vert \bI(\bxi_k+\Delta\bxi) - \bT (\bzero)\Vert}_2^2
\label{eq:lk_iterative}
\end{equation}
%
using first-order Taylor expansion
\begin{equation}
\min_{\Delta\bxi} {\Bigg\Vert\bI(\bxi_k) + \frac{\partial \bI(\bxi_k)}{\partial \bxi} \Delta\bxi - \bT (\bzero)\Bigg\Vert}_2^2
\label{eq:lk_linearized}
\end{equation}
Note that the ``steepest descent image'' $\partial \bI(\bxi_k)/\partial \bxi$
needs to be recomputed at every iteration as it depends on $\bxi_k$.

\boldparagraph{Inverse Compositional Algorithm}
The \textit{inverse compositional} (IC) algorithm \cite{Baker2002} avoids this by applying the warp increments $\Delta \bxi$ to the template instead of the image
%
\begin{equation}
\min_{\Delta\bxi} {\Vert \bI(\bxi_k) - \bT(\Delta\bxi)\Vert}_2^2
\label{eq:ic_iterative}
\end{equation}
using the warp parameter update $\bxi_{k+1} = \bxi_k \circ (\Delta \bxi)^{-1}$.
In the corresponding linearized equation
\begin{equation}
\min_{\Delta\bxi} {\Bigg\Vert\bI(\bxi_k) - \bT (\bzero) - \frac{\partial \bT (\bzero)}{\partial \bxi} \Delta\bxi \Bigg\Vert}_2^2
\label{eq:ic_linearized}
\end{equation}
%
$\partial \bT(\bzero)/\partial \bxi$ does not depend on $\bxi_k$ and can thus be pre-computed, resulting in a more efficient algorithm.


\boldparagraph{Robust M-Estimation}
To handle outliers or ambiguities (\eg, multiple motions), robust estimation \cite{Huber1964AMS,Zhang1997IVC} can be used.
The robust version of the IC algorithm \cite{Baker2003} has the following objective function
%
\begin{equation}
\min_{\Delta\bxi} ~ \br_k(\Delta\bxi)^T\,\bW\,\br_k(\Delta\bxi)
\label{eq:ic_robust}
\end{equation}
where $\br_k(\Delta\bxi) = \bI(\bxi_k) - \bT(\Delta\bxi)$ is the residual between image $\bI$ and template $\bT$ at the $k$'th iteration, and $\bW$ is a diagonal weight matrix that depends on the residual\footnote{We omit this dependency to avoid clutter in the notation.} and is chosen based on the desired robust loss function $\rho$ \cite{Zhang1997IVC}.

\boldparagraph{Optimization}
The minimizer of \eqref{eq:ic_robust} after linearization is obtained as the Gauss-Newton update step \cite{Bjoerck1996}
\begin{equation}
(\bJ^T \bW \bJ) \Delta \bxi = \bJ^T \bW \, \br_k(\bzero)
\label{eq:gauss_newton_robust}
\end{equation}
where
$\bJ=\partial \bT(\bzero)/\partial \bxi$ is the Jacobian of the template $\bT(\bzero)$ with respect to the warp parameters $\bxi$.
As the approximate Hessian $\bJ^T \bW \bJ$ easily becomes ill-conditioned, a damping term is added in practice. This results in the popular Levenberg–Marquardt (trust-region) update equation \cite{Marquardt1963SIAM}:
%
\begin{equation}
\Delta \bxi = {(\bJ^T \bW \bJ + \lambda\,\diag{\bJ^T\bW \bJ})}^{-1} \bJ^T \bW \, \br_k(\bzero)
\label{eq:levenberg_marquardt_robust}
\end{equation}
For different values of $\lambda$, the parameter update $\Delta\bxi$ varies between the Gauss-Newton direction and gradient descent. In practice, $\lambda$ is chosen based on simple heuristics.

\boldparagraph{Limitations}
Despite its widespread utility, the IC method suffers from a number of important limitations.
First, it assumes that the linearized residual leads to an update which iteratively reaches a good local optimum. However, this assumption is invalid in the presence of high-frequency textural information or noise in $\bI$ or $\bT$. 
Second, choosing a good robust loss function $\rho$ is difficult as the true data/residual distribution is often unknown. Moreover, Equation~\eqref{eq:ic_robust} does not capture correlations or higher-order statistics in the inputs $\bI$ and $\bT$ as the residuals operate directly on the pixel values and the weight matrix $\bW$ is diagonal.
Finally, damping heuristics do not fully exploit the information available during optimization and thus lead to suboptimal solutions.

\boldparagraph{Contributions}
In this paper, we propose to combine the best of both (optimization and learning-based) worlds by unrolling the robust IC algorithm into a more general parameterized feed-forward model which is trained end-to-end from data.
In contrast to generic neural network estimators, this allows our algorithm to incorporate knowledge about the structure of the problem (\eg, family of warping functions, 3D geometry) as well as the advantages of a robust iterative estimation framework. At the same time, our approach relaxes the restrictive assumptions made in the original IC formulation \cite{Baker2002} by incorporating trainable modules and learning the entire model end-to-end.

More specifically, we make the following contributions:
\begin{description}
	\item[(A)] We propose a \textbf{Two-View Feature Encoder} which replaces $\bI,\bT$ with feature representations $\bI_\theta,\bT_\theta$ that jointly encode information about the input image $\bI$ and the template $\bT$. This allows our model to exploit spatial as well as temporal correlations in the data.
	\item[(B)] We propose a \textbf{Convolutional M-Estimator} that replaces $\bW$ in \eqref{eq:ic_robust} with a learned weight matrix $\bW_\theta$ which encodes information about $\bI$, $\bT$ and $\br_k$ in a way such that the unrolled optimization algorithm ignores irrelevant or ambiguous information as well as outliers.
	\item[(C)] We propose a \textbf{Trust Region Network} which replaces the damping matrix $\lambda\,\diag{\bJ^T\bW \bJ}$ in \eqref{eq:levenberg_marquardt_robust} with a learned damping matrix $\diag{\blambda_\theta}$ whose diagonal entries $\lambda_\theta$ are estimated from ``residual volumes'' which comprise residuals of a Levenberg-Marquardt update when applying a range of hypothetical $\lambda$ values.
\end{description}
We demonstrate the advantages of combining the classical IC method with deep learning on the task of 3D rigid motion estimation using several challenging RGB-D datasets.
We also provide an extensive ablation study about the relevance of each model component that we propose.
Results on traditional affine 2D image alignment tasks are provided in the supplementary material. Our implementation is publicly accessible.\footnote{\url{https://github.com/lvzhaoyang/DeeperInverseCompositionalAlgorithm}}



\section{Related Work}
\label{sec:related}

We are not the first to inject deep learning into an optimization pipeline.
In this section, we first review classical methods, followed by direct pose regression techniques and related work on learning-based optimization.

\boldparagraph{Classical Methods}
%
%
Direct methods \cite{Horn88IJCV, Irani2000ICCVWORK} that align images using the sum-of-square error objective \eqref{eq:lk_objective} are prone to outliers and varying illuminations.
Classical approaches address this problem by exploiting more robust objective functions \cite{Lucas1981IJCAI,Newcombe2011ICCV}, heuristically chosen patch- \cite{Szeliski2007FTCGV} or gradient-based \cite{Levin2004ECCV} features, and photometric calibration as a pre-processing step \cite{Engel18pami}.
The most common approach is to use robust estimation \eqref{eq:ic_robust} as in \cite{Black1996CVIU}.
However, the selection of a good robust function $\rho$ is challenging and traditional formulations assume that the same function applies to all pixels, ignoring correlations in the inputs.
Moreover, the inversion of the linearized system \eqref{eq:gauss_newton_robust} may still be ill-conditioned \cite{Anandan1989IJCV}. To overcome this problem, soft constraints in the form of damping terms \eqref{eq:levenberg_marquardt_robust} can be added to the objective \cite{Baker2002,Baker2003}. However, this may bias the system to sub-optimal solutions. 

This paper addresses these problems by relaxing the main assumptions of the Inverse Compositional (IC) algorithm \cite{Baker2002,Baker2003} using data-driven learning. More specifically, we propose to learn the feature representation (A), robust estimator (B) and damping (C) jointly to replace the traditional heuristic rules of classical algorithms.


\boldparagraph{Direct Pose Regression}
A notably different approach to classical optimization techniques is to directly learn the entire mapping from the input to the warping parameters $\bxi$ from large amounts of data, spanning from early work using linear hyperplane approximation \cite{Jurie02pami} to recent work using deep neural networks \cite{Ummenhofer2017CVPR, Byravan2017ICRA,Zhou2017CVPRa, Yin2018CVPR, Wang2018CVPRc}.
Prominent examples include single image to camera pose regression
\cite{Kendall2015ICCV, Kendall2017ICCV}, image-based 3D object pose estimation \cite{Manhardt2018ECCV, Sundermeyer2018ECCV} and relative pose 
prediction from two views \cite{Ummenhofer2017CVPR, Byravan2017ICRA}.
However, learning a direct mapping requires high-capacity models and large amounts of training data. Furthermore, obtaining pixel-accurate registrations remains difficult and the learned representations do not generalize well to new domains.


To improve accuracy, recent methods adopt cascaded networks \cite{Wei16cvpr, Ilg17cvpr} and iterative feedback \cite{Kanazawa2018CVPR}. Lin \etal \cite{Lin2017CVPR} combines the multi-step iterative spatial transformer network (STN) with the classical IC algorithm \cite{Baker2002,Baker2003} for aligning 2D images.
Variants of this approach have recently been applied to various 3D tasks:
Li \etal \cite{Li2018ECCV} proposes to iteratively align a 3D CAD model to an image.
Zhou \etal \cite{Zhou2018ECCV} jointly train for depth, pose and optical flow. 

Different from \cite{Lin2017CVPR} and its variants which approximate the pseudo-inverse of the Jacobian implicitly using stacked convolutional layers, we exploit the structure of the optimization problem and explicitly solve the original robust objective \eqref{eq:ic_robust} with learned modules using few parameters.

\boldparagraph{Learning-based Optimization}
Recently, several methods have exploited the differentiable nature of iterative optimization algorithms
by unrolling for a fixed number of iterations. Each iteration is treated as a layer in a neural network \cite{Zheng2015ICCV,Cherabier2018ECCV,Riegler2016ECCV,Kobler2017GCPR,Meinhardt2017ICCV,Ranftl2014GCPR}. In this section, we focus on the most related work which also tackles the least-squares optimization problem \cite{Wang2018CVPRc,Clark18eccv,Ranftl2018ECCV,Tang19iclr}. We remark that most of these techniques can be considered special cases of our more general deep IC framework.


Wang \etal \cite{Wang2018CVPRc} address the 2D image tracking problem by learning an input representation using a two-stream Siamese network for the IC setting.
In contrast to us, they exploit only spatial but not temporal correlations in the inputs (A), leverage a formulation which is not robust (B) and do not exploit trust-region optimization (C).

Clark \etal \cite{Clark18eccv} propose to jointly learn depth and pose estimation by minimizing photometric error using the formulation in \eqref{eq:ic_iterative}. In contrast to us, they do not learn feature representations (A) and neither employ a robust formulation (B) nor trust-region optimization (C).

Ranftl \etal \cite{Ranftl2018ECCV} propose to learn robust weights for \textit{sparse}  feature correspondences and apply their model to fundamental matrix estimation.
As they do not target direct image alignment, their problem setup is different from ours.
Besides, they neither learn input features (A) nor leverage trust-region optimization (C). Instead, they solve their optimization problem using singular value decomposition.

Concurrent to our work, Tang and Tan \cite{Tang19iclr} propose a photometric Bundle Adjustment network by learning to align feature spaces for monocular reconstruction of a static scene. Different from us, they did not exploit temporal correlation in the inputs (A) and do not employ a robust formulation (B). While they propose to learn the damping parameters (C), in contrast to our trust-region volume formulation, they regress the damping parameters from the global average pooled residuals.

 



\begin{figure*}[t!]
	\centering 
	\includegraphics[width=0.82\textwidth]{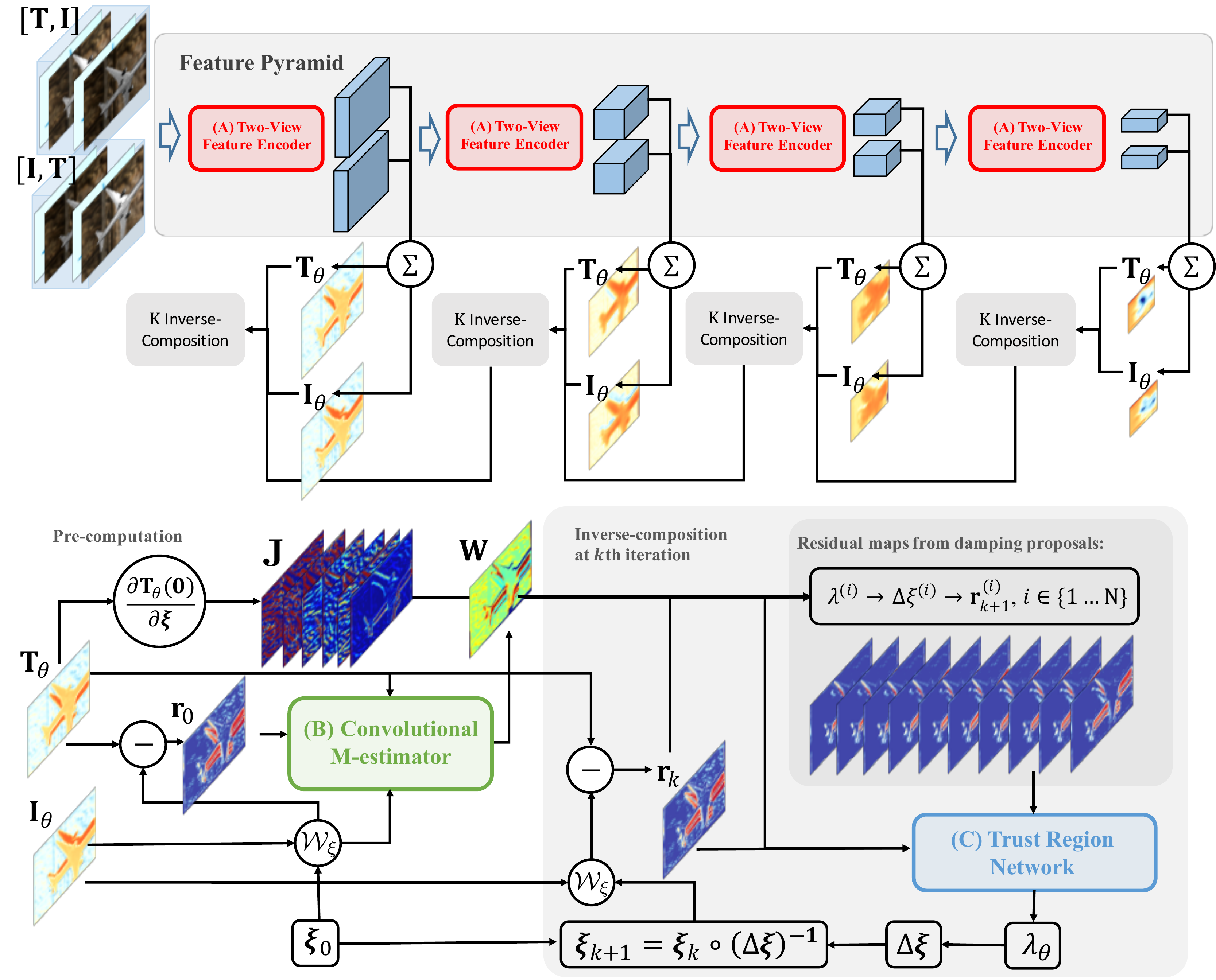} 
	\caption{\textbf{High-level Overview of our Deep Inverse Compositional (IC) Algorithm.}  We stack $[\bI, \bT]$ and $[\bT, \bI]$ as inputs to our \textcolor{red}{\textbf{(A) Two-View Feature Encoder}} pyramid which extracts 1-channel feature maps $\bT_{\theta}$ and $\bI_{\theta}$ at multiple scales using channel-wise summation.
	We then perform $K$ IC steps at each scale using $\bT_{\theta}$ and $\bI_{\theta}$ as input.
	At the beginning of each scale, we pre-compute $\bW$ using our \textcolor{darkgreen}{\textbf{(B) Convolutional M-estimator}}.
	For each of the $K$ IC iterations, we compute the warped image $\bI(\bxi_{k})$ and $\br_{k}$.
	Subsequently, we sample $N$ damping proposals $\lambda^{(i)}$ and compute the proposed residual maps $\br^{(i)}_{k+1}$.
	Our \textcolor{blue}{\textbf{(C) Trust Region Network}} takes these residual maps as input and predicts $\lambda_{\theta}$ for the trust region update step.}
	\label{fig:overview}
	\vspace{-0.4cm}
\end{figure*}

\section{Method}
\label{sec:method}

This section describes our model. A high-level overview over the proposed unrolled inverse-compositional algorithm is given in \figref{fig:overview}.
Using the same notation as in \secref{sec:introduction}, our goal is to minimize the error
by warping the image towards the template, similar to \eqref{eq:ic_robust}:
\begin{equation}
\min_{\bxi} ~ \br(\bxi)^T\,\bW_\theta\,\br(\bxi)
\label{eq:dic_objective}
\end{equation}
The difference to \eqref{eq:ic_robust} is that in our formulation, the weight matrix $\bW_\theta$ as well as the template $\bT_\theta(\bxi)$ and the image $\bI_\theta(\bxi)$ (and thus also the residual $\br(\bxi) = \bI_\theta(\bxi) - \bT_\theta(\bzero)$) depend on the parameters of a learned model $\theta$.

We exploit the inverse compositional algorithm to solve the non-linear optimization problem in \eqref{eq:dic_objective}, \ie, we linearize the objective and iteratively update the warp parameters $\bxi$:
\begin{eqnarray}
\bxi_{k+1} &=& \bxi_k \circ (\Delta \bxi)^{-1}\label{eq:dic_update_xi}\\
\Delta \bxi &=& {(\bJ^T \bW_\theta \bJ + \diag{\blambda_\theta})}^{-1} \bJ^T \bW_\theta \, \br_k\qquad\label{eq:dic_update_delta_xi}\\
\br_k &=& \bI_{\theta}(\bxi_k) - \bT_\theta(\bzero)\label{eq:dic_update_residual}\\
\bJ&=&\partial \bT_{\theta}(\bzero)/\partial \bxi\label{eq:dic_update_jacobian}
\label{eq:dic_update}
\end{eqnarray}
starting from $\bxi_0=\bzero$.

The most notable change from \eqref{eq:levenberg_marquardt_robust} is that the image features ($\bI_\theta,\bT_\theta$), the weight matrix ($\bW_\theta$) and the damping factors $\blambda_\theta$ are predicted by learned functions which have been collectively parameterized by $\theta=\{\theta_{\bI},\theta_{\bW},\theta_{\blambda}\}$. Note that also the residual~$\br_k$ as well as the Jacobian~$\bJ$ implicitly depend on the parameters $\theta$, though we have omitted this dependence here for notational clarity.

We will now provide details about these mappings.

\boldparagraph{(A) Two-View Feature Encoder}
We use a fully convolutional neural network $\phi_\theta$ to extract feature maps from the image $\bI$ and the template $\bT$:
\begin{eqnarray}
\bI_\theta&=&\phi_\theta([\bI,\bT])\\
\bT_\theta&=&\phi_\theta([\bT,\bI])
\end{eqnarray}
Here, the operator $[\cdot,\cdot]$ indicates concatenation along the feature dimension.
Instead of $\bI$ and $\bT$ we then feed $\bI_\theta$ and $\bT_\theta$ to the residuals in \eqref{eq:dic_update_residual} and to the Jacobian in \eqref{eq:dic_update_jacobian}. Note that we use the notation $\bI_\theta(\bxi_k)$ in \eqref{eq:dic_update_residual} to denote that the feature map $\bI_\theta$ is warped by a warping function that is parameterized via $\bxi$. More formally, this can be stated as $\bI_\theta(\bxi)=\bI_\theta(\cW_{\bxi}(\bx))$, where $\bx\in\nR^2$ denote pixel locations and $\cW_{\bxi}:\nR^2 \rightarrow \nR^2$ is a warping function that maps a pixel location to another pixel location. For instance, $\cW_{\bxi}$ may represent the space of 2D translations or affine transformations. In our experiments, we will focus on challenging 3D rigid body motions, using RGB-D inputs and representing $\bxi$ as an element of the Euclidean group $\bxi\in\se{3}$.

Note that compared to directly using the image $\bI$ and $\bT$ as input, our features capture high-order \textit{spatial} correlations in the data, depending on the receptive field size of the convolutional network. Moreover, they also capture \textit{temporal} information as they operate on both $\bI$ and $\bT$ as input.

\boldparagraph{(B) Convolutional M-Estimator}
We parameterize the weight matrix $\bW_\theta$ as a diagonal matrix whose elements are determined by a fully convolutional network $\psi_\theta$ that operates on the feature maps and the residual:
\begin{equation}
\bW_\theta=\diag{\psi_\theta(\bI_\theta(\bxi_k),\bT_\theta(\bzero),\br_k)}
\end{equation}
%
as is illustrated in \figref{fig:overview}.
Note that this enables our algorithm to reason about relevant image information while capturing spatial-temporal correlations in the inputs which is not possible with classical M-Estimators. Furthermore, we do not restrict the implicit robust function $\rho$ to a particular error model, but instead condition $\rho$ itself on the input. This allows for learning more expressive noise models.

\boldparagraph{(C) Trust Region Network}
For estimating the damping $\blambda_\theta$ we use a fully-connected network as illustrated in \figref{fig:overview}. We first sample a set of scalar damping proposals $\lambda_i$ on a logarithmic scale and compute the resulting Levenberg-Marquardt update step as
\begin{equation}
\Delta \bxi_i = {(\bJ^T \bW \bJ + \lambda_i\,\diag{\bJ^T\bW \bJ})}^{-1} \bJ^T \bW \, \br_k(\bzero)
\end{equation}
We stack the resulting $N$ residual maps
\begin{equation}
\br_{k+1}^{(i)} = \bI_{\theta}(\bxi_k \circ (\Delta \bxi_i)^{-1}) - \bT_\theta(\bzero)
\end{equation}
into a single feature map, flatten it, and pass it to a fully connected neural network $\nu_\theta$ that outputs the damping parameters $\blambda_\theta$:
\begin{equation}
\blambda_\theta = \nu_\theta\left(\bJ^{T}\bW\bJ, \left[\bJ^{T}\bW\br_{k+1}^{(1)},\dots,\bJ^{T}\bW\br_{k+1}^{(N)}\right]\right)
\end{equation}
%
The intuition behind our trust region networks is that the residuals predicted using the Levenberg-Marquardt update comprise valuable information about the damping parameter itself. This is empirically confirmed by our experiments.

%

\boldparagraph{Coarse-to-Fine Estimation}
To handle large motions, it is common practice to apply direct methods in a coarse-to-fine fashion. We apply our algorithm at four pyramid levels with three iterations each. Our entire model including coarse-to-fine estimation is illustrated in \figref{fig:overview}. We extract features at all four scales using a single convolutional neural network with spatial average pooling between pyramid levels. We start with $\bxi=\bzero$ at the coarsest pyramid level, perform 3 iterations of our deep IC algorithm, and proceed with the next level until we reach the original image resolution.

\boldparagraph{Training and Inference}
For training and inference, we unroll the iterative algorithm in equations \eqref{eq:dic_update_xi}-\eqref{eq:dic_update_jacobian}. We obtain the gradients of the resulting computation graph using auto-differentiation.
Details about the network architectures which we use can be found in the supplementary material.

\section{Experiments}
\label{sec:experiments}

We perform our experiments on the challenging task of 3D rigid body motion estimation
from RGB-D inputs\footnote{Additional experiments on classical affine 2D motion estimation tasks can be found in the supplementary material.}.
Apart from the simple scenario of purely static scenes where only the camera is moving, we also consider scenes where both the camera as well as objects are in motion, hence resulting in strong ambiguities.
We cast this as a supervised learning problem: using the ground truth motion, we train the models to resolve these ambiguities by learning to focus either on the foreground or the background.



\boldparagraph{Warping Function}
Given pixel $\bx \in \nR^{2}$, camera intrinsics~$\bK$ and depth $D(\bx)$, we define the warping $\warp_{\bxi}(\bx)$ induced by rigid body transform $\bT_{\bxi}$ with $\bxi \in \se{3}$ as
\begin{equation}\label{eq:warp_function}
\warp_{\bxi}(\bx) = \, \bK \, \bT_{\bxi} \, D(\bx)  \bK^{-1} \, \bx
\end{equation}
Using the warped coordinates, compute $\bI_{\theta}({\bxi})$ via bilinear sampling from $\bI_{\theta}$ and set the warped feature value to zero for all occluded areas (estimated via z-buffering).

\boldparagraph{Training Objective}
To balance the influences of translation and rotation we follow \cite{Li2018ECCV} and exploit the 3D End-Point-Error (EPE) as loss function. Let $\bp=D(\bx) \bK^{-1} \bx$ denote the 3D point corresponding to pixel $\bx$ in image $\bI$ and let $\cP$ denote the set of all such 3D points. We minimize the following loss function
%
\begin{equation}\label{eq:train_loss}
\nL = \frac{1}{|\cP|}\sum_{l \in \cL}\sum_{\bp \in \cP}\norm{\bT_{gt}\,\bp - \bT(\bxi_{l})\,\bp}^{2}_2
\end{equation}
where $\cL$ denotes the set of coarse-to-fine pyramid levels (we apply our loss at the final iteration of every pyramid level)
and $\bT_{gt}$ is the ground truth transformation. 

\boldparagraph{Implementation}
We use the Sobel operator to compute the gradients in $\bT$ and analytically derive $\bJ$. We calculate the matrix inverse on the CPU since we observed that inverting a small dense matrices $\hessian \in \nR^{6\times6}$ is significantly faster on the CPU than on the GPU. In all our experiments we use $N=10$ damping proposals for our Trust Region Network, sampled uniformly in logscale between $[10^{-5},10^{5}]$. We use four coarse-to-fine pyramid levels with three iterations each.
We implemented our model and the baselines in PyTorch. All experiments start with a fixed learning rate of 0.0005 using ADAM \cite{Kingma2015ICLR}. We train a total of 30 epochs, reducing the learning rate at epoch [5,10,15]. 

\subsection{Datasets}

We systematically train and evaluate our method on four datasets 
which we now briefly describe.

\boldparagraph{MovingObjects3D} For the purpose of systematically evaluating highly varying object motions, we
downloaded six categories of 3D models from ShapeNet \cite{Chang15arXiv}. For each object category, we rendered 200 video sequences with 100 frames in each sequence using Blender. We use data rendered from the categories 'boat' and 'motorbike' as test set and data from categories 'aeroplane', 'bicycle', 'bus', 'car' as training set. From the training set we use the first 95\% of the videos for training and the remaining 5\% for validation. In total, we obtain 75K images for training, 5K images for validation, and 25K for testing. We further subsample the sequences using sampling intervals $\{1,2,4\}$ in order to obtain small, medium and large motion subsets.

For each rendered sequence, we randomly select one 3D object model within the chosen category and stage it in a static 3D cuboid room with random wall textures and four point light sources. We randomly choose the camera viewpoint, point light source position and object trajectory, see supplementary material for details.
This ensures diversity in object motions, textures and illumination. The videos also contain frames where the object is only partially visible. We exclude all frames where the entire object is not visible.

\boldparagraph{BundleFusion} To evaluate camera motion estimation in a static environment, we use the eight publicly released scenes from BundleFusion\footnote{\url{http://graphics.stanford.edu/projects/bundlefusion/}} \cite{Dai17tog} which provide fully synchronized RGB-D sequences. We hold out 'copyroom' and 'office2' scenes for test and split the remaining scenes into training (first 95\% of each trajectory) and validation (last 5\%). We use the released camera trajectories as ground truth. We subsampled frames at intervals $\{2,4,8\}$ to increase motion magnitudes and hence the level of difficulty.


\boldparagraph{DynamicBundleFusion}
To further evaluate camera motion estimation under heavy occlusion and motion ambiguity, we use the DynamicBundleFusion dataset \cite{Lv18eccv} which augments the scenes from BundleFusion with non-rigidly moving human subjects as distractors. We use the same training, validation and test split as above. We train and evaluate frames subsampled at intervals $\{1,2,5\}$ due to the increased difficulty of this task.

\boldparagraph{TUM RGB-D SLAM} 
We evaluate our camera motion estimates on the TUM RGB-D SLAM dataset \cite{Sturm12iros}. We hold out 'fr1/360', 'fr1/desk', 'fr2/desk' and 'fr2/pioneer\_360' for testing and split the remaining trajectories into training (first 95\% of each trajectory) and validation (last 5\%). We randomly sample frame intervals from \{1,2,4,8\}. All images are resized to $160\times120$ pixels and depth values outside the range [0.5m, 5.0m] are considered invalid.


\begin{table*}[t!]
	\centering
	\scriptsize
\begin{tabular}{@{}llcccccccccc@{}}
	\toprule
	& \multicolumn{1}{c}{\multirow{3}{*}{Model Descriptions}} 
	& \multicolumn{3}{c}{3D EPE (cm) $\downarrow$ on Validation/Test }    
	& \multicolumn{3}{c}{$\eulerangle$ (Deg) $\downarrow$ / $\translation$ (cm) $\downarrow$ / ($\translation < 5$ (cm) \& $\eulerangle < 5^{\circ}$) $\uparrow$ on Test}  
	\\ \cmidrule(lr){3-8}
	& & Small & Medium & Large & Small & Medium & Large \\ \midrule
	\multirow{4}{*}{\rotatebox[origin=c]{90}{ICP}} 
	& Point-Plane ICP \cite{Chen92ivc} & 4.88/4.28 & 10.13/8.74 & 20.24/17.43 & 4.32/10.54/66.23\% & 8.29/20.90/33.4\% & 15.40/40.45/11.2\% \\
	& Point-Point ICP \cite{Besl92pami} & 5.02/4.38 & 10.33/9.06 & 20.43/17.68 &4.04/10.51/70.0\% & 7.89/20.15/33.8\% & 15.33/40.52/11.1\% \\
	& Oracle Point-Plane ICP \cite{Chen92ivc} & 3.91/3.31 & 10.68/9.63 & 22.53/19.98 & \textbf{2.74}/9.75/\textbf{78.6\%} & 8.31/19.72/\textbf{40.3\%} & 16.64/41.40/\textbf{16.4\%}  \\
	& Oracle Point-Point ICP \cite{Besl92pami} & 4.34/3.99 & 11.83/10.29 & 21.27/26.13 & 3.81/10.11/75.0\% & 9.30/26.41/39.1\% & 19.53/62.30/14.1\% \\ \midrule
	\multirow{3}{*}{\rotatebox[origin=c]{90}{DPR}}         
	& PoseCNN & 5.18/4.60 & 10.43/9.20 & 20.08/17.74 & 3.91/10.51/69.8\% & 7.89/21.10/34.7\% & 15.34/40.54/11.0\% \\ 
	& IC-PoseCNN & 5.14/4.56 & 10.40/9.13 & 19.80/17.31 & 3.93/10.49/70.1\% & 7.90/21.01/34.8\% & 15.31/40.51/11.1\% \\
	& Cascaded-PoseCNN & 5.24/4.68 & 10.43/9.21 & 20.32/17.30 & 3.90/10.50/70.0\% & 7.90/21.11/34.6\% & 15.32/40.60/10.8\%  \\ \midrule
	\multirow{7}{*}{\rotatebox[origin=c]{90}{Learning Optim}}  
	& No learning & 11.66/11.26  & 21.85/22.95 & 37.01/38.88 & 4.29/15.90/66.8\% & 8.33/31.80/32.2\% & 15.78/55.70/10.44\% \\
	& IC-FC-LS-Net, adapted from \cite{Clark18eccv} & 4.96/4.62 & 10.49/9.21 &  20.31/17.34 &  3.94/10.49/70.0\%  & 7.90/21.20/34.5\% & 15.33/40.63/11.2\%  \\
	& DeepLK-6DoF, adapted from \cite{Wang18icra} & 4.41/3.75 & 9.05/7.54 & 18.46/15.33 & 4.03/10.25/68.1\% &  7.96/20.34/33.6\% &  15.43/39.56/10.6\% \\
	& Ours: (A)	& 4.35/3.66 & 8.80/7.23 & 18.28/15.06 & 4.09/10.19/68.6\% & 8.00/20.28/32.9\% &  15.37/39.49/10.9\%  \\
	& Ours: (A)+(B)	& 4.33/3.26 & 8.84/7.30 & 18.14/15.04 & 4.02/10.11/69.2\% & 7.96/20.26/33.4\% & 15.35/39.56/10.9\%  \\
	& Ours: (A)+(B)+(C) & \textbf{3.58}/2.91 & \textbf{7.30}/\textbf{5.94} &  \textbf{15.48}/\textbf{12.96} & 3.74/\textbf{9.73}/74.5\% & 7.41/\textbf{19.60}/38.2\% & 14.71/\textbf{38.39}/12.9\%   \\
	& Ours: (A)+(B)+(C) (No WS) & 3.62/\textbf{2.89} & 7.54/6.08 & 16.00/12.98 & 3.69/\textbf{9.73}/74.9\% & \textbf{7.37}/19.74/38.4\% & \textbf{14.65}/38.69/12.8\% \\
	& Ours: (A)+(B)+(C) ($K=1$) & 4.12/3.37 & 8.64/7.08 & 17.67/14.92 & 3.85/9.95/71.2\% & 7.80/20.13/35.5\% & 15.21/39.30/11.6\%  \\
	& Ours: (A)+(B)+(C) ($K=5$) & 3.60/2.92 & 7.49/6.09 & 16.06/13.01 & 3.68/9.77/74.4\% & 7.46/19.69/37.8\% & 14.76/38.65/12.4\% \\
	\bottomrule
\end{tabular}

\caption{\textbf{Quantitative Evaluation on MovingObjects3D.} We evaluate the average  3D EPE, angular error in $\eulerangle$ (Euler angles), translation error $\bt$ and success ratios $\bt<5$ \& $\eulerangle<5^{\circ}$ for three different motion magnitudes \{Small, Medium, Large\} which correspond to frames sampled from the original videos using frame intervals $\{1,2,4\}$.}
\label{tab:synthetic_quantitative}
\end{table*}

\begin{table*}[]
	\centering
	\scriptsize
\setlength{\tabcolsep}{6.8pt}
\begin{tabular}{@{}llcccccccccc@{}}
	\toprule
	& \multicolumn{1}{c}{\multirow{2}{*}{Model Descriptions}} 
	& \multicolumn{3}{c}{\begin{tabular}[c]{@{}c@{}}3D EPE (cm) $\downarrow$ Validation/Test \\ on BundleFusion\cite{Dai17tog} \end{tabular}  }    
	& \multicolumn{3}{c}{\begin{tabular}[c]{@{}c@{}}3D EPE (cm) $\downarrow$  Validation/Test \\ on DynamicBundleFusion \cite{Lv18eccv} \end{tabular} } & \multirow{2}{*}{\begin{tabular}[c]{@{}c@{}}Model \\ Size (K) \end{tabular}} & \multirow{2}{*}{\begin{tabular}[c]{@{}c@{}}Inference \\ Time (ms) \end{tabular}} \\ \cmidrule(lr){3-8}
	&  & Small & Medium  & Large  & Small & Medium & Large & &  \\ \midrule
	\multirow{2}{*}{\rotatebox[origin=c]{90}{ICP}} 
	& Point-Plane ICP \cite{Chen92ivc} & 2.81/2.01 & 6.85/4.52 & 16.20/11.11 & 1.26/0.97 & 2.64/2.09 & 10.38/4.89 & - & 310 \\
	& Point-Point ICP \cite{Besl92pami} & 3.62/2.48 & 8.17/5.72 & 17.23/12.58 & 1.42/1.08 & 3.40/2.42 & 13.09/7.43 & - & 402 \\ \midrule
	\multirow{3}{*}{\rotatebox[origin=c]{90}{DPR}}         
	& PoseCNN & 4.76/3.41 & 9.78/6.85 & 17.90/13.31 & 2.20/1.65 & 4.22/3.19 & 13.90/8.24 & 19544 &  5.7 \\ 
	& IC-PoseCNN & 4.32/3.26 & 8.98/6.52 & 16.30/12.81 & 2.21/1.66 & 4.24/3.18 & 12.99/8.05  & 19544 & 14.1 \\
	& Cascaded-PoseCNN & 4.46/3.41 & 9.38/6.81 & 16.50/13.02 & 2.20/1.60 & 4.13/3.15 & 12.97/8.15 & 58632 & 14.1 \\ \midrule
	\multirow{7}{*}{\rotatebox[origin=c]{90}{Learning Optim}}  
	& No learning &  4.52/3.35 &  8.64/6.30  &  17.06/12.51  &  4.89/3.39  & 5.64/4.69  &  13.88/8.58  & - & 1.5  \\
	& IC-FC-LS-Net, adapted from \cite{Clark18eccv}  &  4.04/3.03 & 9.06/6.85 & 17.15/13.32 & 2.33/1.80 & 4.43/3.45 &  13.84/8.35 &  674 & 1.7 \\
	& DeepLK-6DoF, adapted from \cite{Wang18icra} & 4.09/2.99  & 8.14/5.84 & 16.81/12.27 & 2.15/1.72 & 3.78/3.12 & 12.73/7.22 & 596 & 2.2 \\
	& Ours: (A) 	& 3.59/2.65 & 7.68/5.46 & 16.56/11.92 &  2.01/1.65 & 3.68/2.96 &  12.68/7.11 &  597 & 2.2 \\
	& Ours: (A)+(B) &  2.42/1.75 & 5.39/3.47 & 12.59/8.40 & 2.22/1.70 & 3.61/2.97  &  12.57/6.88  & 622 & 2.4 \\
	& Ours: (A)+(B)+(C) & 2.27/\textbf{1.48} & \textbf{5.11}/\textbf{3.09} & \textbf{12.16}/7.84 & 1.09/0.74 & 2.15/1.54 &  9.78/4.64 & 662 & 7.6  \\
	& Ours: (A)+(B)+(C) (No WS) & \textbf{2.14}/1.52 & 5.15/3.10 & 12.26/\textbf{7.8}1  & \textbf{0.93}/\textbf{0.61} & \textbf{1.86}/\textbf{1.32} & \textbf{8.88}/\textbf{3.82}  & 883 & 7.6 \\
	\bottomrule
\end{tabular}
\caption{\textbf{Quantitative Evaluation on BundleFusion and DynamicBundleFusion.} In BundleFusion \cite{Dai17tog}, the motion magnitudes \{Small, Medium, Large\} correspond to frame intervals $\{2,4,8\}$. In DynamicBundleFusion \cite{Lv18eccv}, the motion magnitudes \{Small, Medium, Large\} correspond to frame intervals $\{1,2,5\}$ (we reduce the intervals due to the increased difficulty). }
\label{tab:bundlefusion_quantitative}
\vspace{-0.3cm}
\end{table*}

\begin{table}[t!]
	\centering
	\scriptsize
	\setlength{\tabcolsep}{4pt}
	\begin{tabular}{@{}llccccc@{}}
		\toprule
		& \multicolumn{1}{c}{\multirow{2}{*}{}} 
		& \multicolumn{4}{c}{\begin{tabular}[c]{@{}c@{}} mRPE: $\theta$ (Deg) $\downarrow$ / $t$ (cm) $\downarrow$  \end{tabular}  }    \\
		& & KF 1  & KF 2 &  KF 4 & KF 8 \\ \midrule
		& RGBD VO \cite{Steinbrucker11iccvw} &  0.55/1.03 & 1.39/2.81 & 3.99/5.95 & 9.20/13.83   \\
		& Ours: (A) & 0.53/1.17 & 0.97/2.63 & 2.87/6.89 & 7.63/12.16 \\
		& Ours: (A)+(B) & 0.51/1.14 & 0.87/2.44 & 2.60/6.56 & 7.30/11.21 \\
		& Ours: (A)+(B)+(C) & \textbf{0.45}/\textbf{0.69}  & \textbf{0.63}/\textbf{1.14} & \textbf{1.10}/\textbf{2.09} & \textbf{3.76}/\textbf{5.88} \\
		\bottomrule
	\end{tabular}
	\caption{\textbf{Results on TUM RGB-D Dataset [45].} This table shows the mean relative pose error (mRPE) on our test split of the TUM RGB-D Dataset [45]. KF denotes the size of the key frame intervals. Please refer to the supplementary materials for a detailed evaluation of individual trajectories.}
	\label{tab:tum_quantitative}
	\vspace{-0.3cm}
\end{table}

\begin{figure*}[t!]
	\centering 
\begin{tabularx}{\textwidth}{cccccc}
\vspace{0.03cm}
$\bT$ & 
\parbox[c]{0.15\textwidth}{\includegraphics[width=0.18\textwidth]{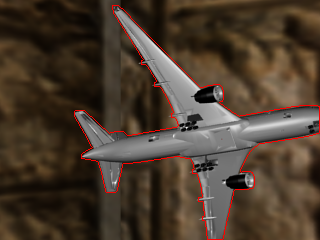}}&
\parbox[c]{0.15\textwidth}{\includegraphics[width=0.18\textwidth]{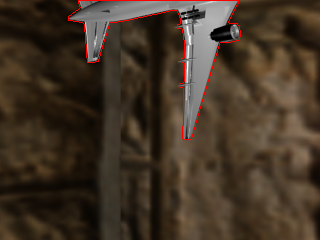}}	 &
\parbox[c]{0.15\textwidth}{\includegraphics[width=0.18\textwidth]{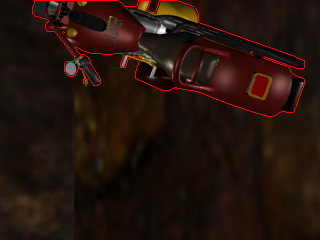}}	 &
\parbox[c]{0.15\textwidth}{\includegraphics[width=0.18\textwidth]{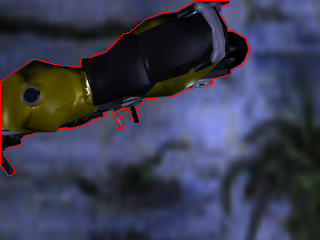}}	 &	
\parbox[c]{0.15\textwidth}{\includegraphics[width=0.18\textwidth]{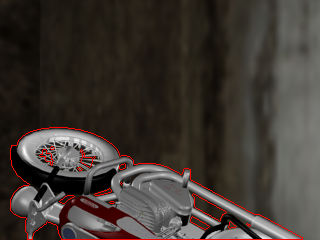}} \\ \vspace{0.03cm}
$\bI$ & 
\parbox[c]{0.15\textwidth}{\includegraphics[width=0.18\textwidth]{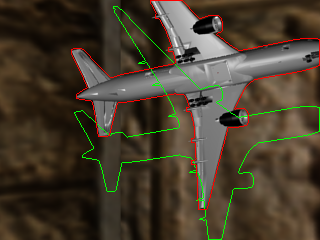}}&
\parbox[c]{0.15\textwidth}{\includegraphics[width=0.18\textwidth]{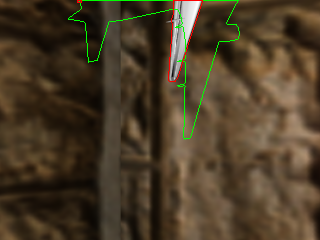}}	 &
\parbox[c]{0.15\textwidth}{\includegraphics[width=0.18\textwidth]{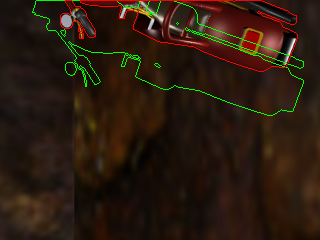}}	 &
\parbox[c]{0.15\textwidth}{\includegraphics[width=0.18\textwidth]{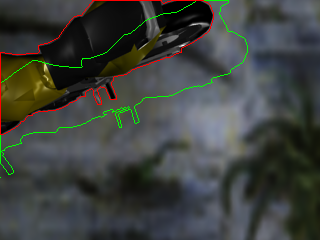}}	 &	
\parbox[c]{0.15\textwidth}{\includegraphics[width=0.18\textwidth]{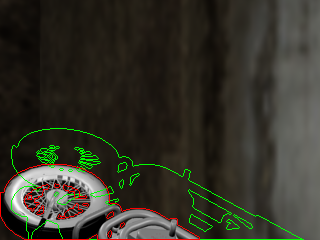}} \\ \vspace{0.03cm}
$\bI(\bxi^{\text{GT}})$ & 
\parbox[c]{0.15\textwidth}{\includegraphics[width=0.18\textwidth]{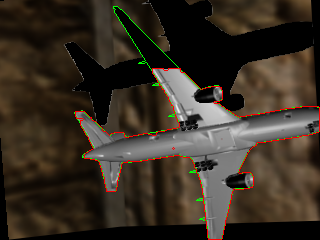}}&
\parbox[c]{0.15\textwidth}{\includegraphics[width=0.18\textwidth]{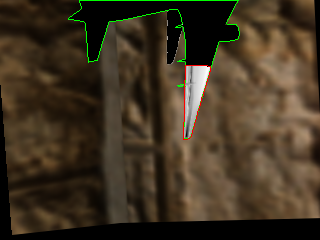}}	 &
\parbox[c]{0.15\textwidth}{\includegraphics[width=0.18\textwidth]{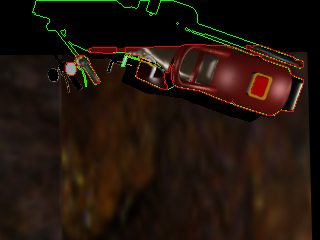}}	 &
\parbox[c]{0.15\textwidth}{\includegraphics[width=0.18\textwidth]{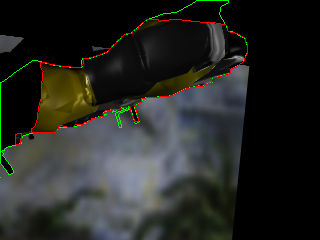}}	 &	
\parbox[c]{0.15\textwidth}{\includegraphics[width=0.18\textwidth]{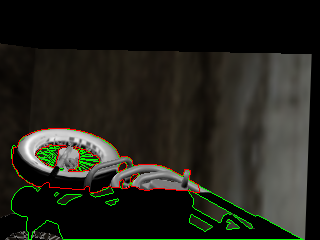}} \\ \vspace{0.03cm}
$\bI(\bxi^{\star})$ & 
\parbox[c]{0.15\textwidth}{\includegraphics[width=0.18\textwidth]{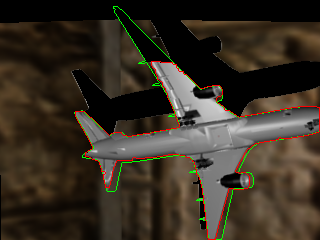}}&
\parbox[c]{0.15\textwidth}{\includegraphics[width=0.18\textwidth]{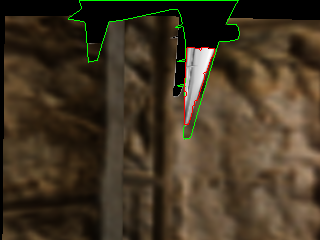}}	 &
\parbox[c]{0.15\textwidth}{\includegraphics[width=0.18\textwidth]{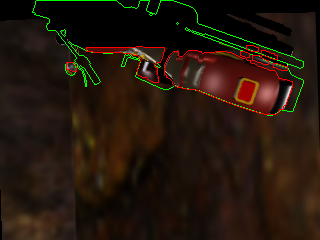}}	 &
\parbox[c]{0.15\textwidth}{\includegraphics[width=0.18\textwidth]{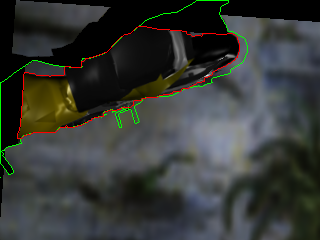}}	 &	
\parbox[c]{0.15\textwidth}{\includegraphics[width=0.18\textwidth]{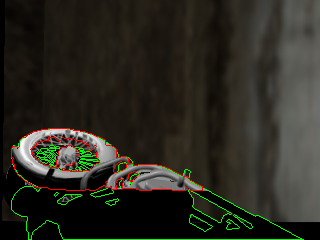}} \\
\end{tabularx}
\caption{\textbf{Qualitative Results  on MovingObjects3D.} Visualization of the warped image $\bI(\bxi)$ using the ground truth object motion $\bxi^{\text{GT}}$ (third row) and the object motion $\bxi^{\star}$ estimated using our method (last row) on the MovingObjects3D \emph{validation} (left two) and \emph{test} sets (right three). In $\bI$, we plot the instance boundary in \textcolor{red}{red} and the instance boundary of $\bT$ in \textcolor{darkgreen}{green} as comparison. Note the difficulty of the task (truncation, independent background object) and the high quality of our alignments. Black regions in the warped image are due to truncation or occlusion.}
\label{fig:qualitative_comparison}
\vspace{-0.3cm}
\end{figure*}

\subsection{Baselines} 

\noindent We implemented the following baselines.

\boldparagraph{ICP} 
We use classical \emph{Point-to-Plane ICP} \cite{Chen92ivc} and \emph{Point-to-Point-ICP}\cite{Besl92pami} implemented in Open3D \cite{Zhou18arXiv}. To examine the effect of ambiguity between the foreground and background in the object motion estimation task, we also evaluate a version for which we provide the ground truth instance segmentation to both methods. Note that this is an upper bound to the performance achievable by ICP methods. We thus call these the \emph{Oracle} ICP methods.

\boldparagraph{RGB-D Visual Odometry} 
We compare to the RGB-D visual odometry method \cite{Steinbrucker11iccvw} implemented in Open3D \cite{Zhou18arXiv} on TUM RGBD SLAM datasets for visual odometry.

\boldparagraph{Direct Pose Regression} We compare three different variants that directly predict the mapping $f:\bI,\bT \rightarrow \bxi$. All three networks use Conv1-6 encoder layers from FlowNet-Simple\cite{Dosovitskiy2015ICCV} as two-view regression backbones. We use spatial average pooling after the last feature layer followed by a fully-connected layer to regress $\bxi$. All three CNN baselines are trained using the loss in \eqref{eq:train_loss}.
\begin{itemize}[noitemsep,nolistsep,leftmargin=*]
	\item \textbf{PoseCNN}: A feed-forward CNN that directly predicts $\bxi$.
	\item \textbf{IC-PoseCNN}: A PoseCNN with iterative refinement using the IC algorithm, similar to \cite{Lin2017CVPR} and \cite{Li2018ECCV}. We noticed that training becomes unstable and performance saturates when increasing the number of iterations. For all our experiments, we thus used three iterations.
	\item \textbf{Cascaded-PoseCNN}: A cascaded network with three iterations, similar to IC-PoseCNN but with independent weights for each iteration.
\end{itemize}
	
\boldparagraph{Learning-based Optimization}
We implemented the following related algorithms within our deep IC framework. For all methods, we use the same number of iterations, training loss and learning rate as used for our method.  
\begin{itemize}[noitemsep,nolistsep,leftmargin=*]
	\item \textbf{DeepLK-6DoF}: We implemented a variant of DeepLK \cite{Wang18icra} which predicts the 3D transformation $\bxi \in \se{3}$ instead of translation and scale prediction in their original 2D task. We use Gauss-Newton as the default optimization for this approach and no Convolutional M-Estimator. A comparison of this approach with our method when using only the two-view feature network (A) shows the benefits of our two-view feature encoder.
	\item \textbf{IC-FC-LS-Net}: We also implemented LS-Net \cite{Clark18eccv} within our IC framework with the following differences to the original paper. First, we do not estimate or refine depth. Second, we do not use a separate network to provide an initial pose estimation. Third, we replace their LSTM layers with three fully connected (FC) layers which take the flattened $\bJ^{T}\bW\bJ$ and $\bJ^{T}\bW\br_{k}$ as input.
\end{itemize}

\boldparagraph{Ablation Study}
We use (A), (B), (C) to refer to our contributions in Sec.\ref{sec:introduction}. We set $\bW$ to the identity matrix when the Convolutional M-Estimator (B) is not used and use Gauss-Newton optimization in the absence of the Trust Region Network (C). We consider the following configurations:
\begin{itemize}[noitemsep,nolistsep,leftmargin=*]
\item \textbf{Ours (A)+(B)+(C):} Our proposed method with shared weights. We perform coarse-to-fine iterations on four pyramid levels with three IC iterations at each level. We use shared weights for all iterations in (B) and (C). 
\item \textbf{Ours (A)+(B)+(C) (No WS):} A version of our method without shared weights. All settings are the same as above except that the network for (B) and (C) have independent weight parameters at each coarse-to-fine scale. 
\item \textbf{Ours (A)+(B)+(C) ($K$ iterations/scale):} The same network as the default setting with shared weights, except that we change the inner iteration number $K$. 
\item \textbf{No Learning:} Vanilla coarse-to-fine IC alignment minimizing photometric error \eqref{eq:ic_iterative} without learned modules.
\end{itemize}


\subsection{Results and Discussion}  
\label{sec:egocentric_estimation}

\tabref{tab:synthetic_quantitative} and \tabref{tab:bundlefusion_quantitative} summarize our main results.
For each dataset, we evaluate the method separately for three different motion magnitudes \{Small, Medium, Large\}. In \tabref{tab:synthetic_quantitative}, \{Small, Medium, Large\} correspond to frames sampled from the original videos at intervals $\{1,2,4\}$. In \tabref{tab:bundlefusion_quantitative}, [Small, Medium, Large] correspond to frame intervals $\{2,4,8\}$ on BundleFusion and $\{1,2,5\}$ on DynamicBundleFusion. We show the following metrics/statistics:
%
\begin{itemize}[noitemsep,nolistsep,leftmargin=*]
	\item \textbf{3D End-Point-Error (3D EPE):} This metric is defined in \eqref{eq:train_loss}. We only evaluate errors on the rigidly moving objects, \ie, the moving objects in MovingObjects3D and the rigid background mask in DynamicBundleFusion.
	\item \textbf{Object rotation and translation:} We evaluate 3D rotation using the norm of Euler angles $\eulerangle$, translation $\bt$ in cm and the success ratio ($\translation < 5$ (cm) \& $\eulerangle < 5^{\circ}$), in \tabref{tab:synthetic_quantitative}.
	\item \textbf{Relative Pose Error}: We follow the TUM RGBD metric \cite{Sturm12iros} using relative axis angle $\theta$ and translation $\bt$ in cm. 
	\item \textbf{Model Weight:} The number of learnable parameters. 
	\item \textbf{Inference speed:} The forward execution time for an image-pair of size $160\times120$, using GTX 1080 Ti.
\end{itemize}
\boldparagraph{Comparison to baseline methods}
Compared to all baseline methods (\tabref{tab:synthetic_quantitative} row 1-10 and \tabref{tab:bundlefusion_quantitative} row 1-8), our full model ((A)+(B)+(C)) achieves the overall best performance across different motion magnitudes and datasets while maintaining fast inference speed. 
Compared to ICP methods (ICP rows in \tabref{tab:synthetic_quantitative} and \tabref{tab:bundlefusion_quantitative}) and classical method (No Learning in \tabref{tab:synthetic_quantitative} and \tabref{tab:bundlefusion_quantitative}), our method achieves better performance without instance information at runtime.
Compared to RGB-D visual odometry \cite{Steinbrucker11iccvw}, our method works particularly well in unseen scenarios and in the presence of large motions (\tabref{tab:tum_quantitative}).
Besides, note that our model can achieve better performance with a significantly smaller number of weight parameters compared to direct image-to-pose regression (DRP rows in \tabref{tab:synthetic_quantitative} and \tabref{tab:bundlefusion_quantitative}).  \figref{fig:qualitative_comparison} shows a qualitative comparison of our method on the MovingObject3D dataset by visualizing $\bI(\bxi)$. Note that our method achieves excellent two-view image alignments, despite large motion, heavy occlusion and varying illumination of the scene. 

\boldparagraph{Ablation Discussion}
Across all ablation variants and datasets, our model achieves the best performance by combining all three proposed modules ((A)+(B)+(C)). This demonstrates that all components are relevant for robust learning-based optimization.
Note that the influence of the proposed modules may vary according to the properties of the specific task and dataset. For example, in the presence of noisy depth estimates from real data, learning a robust M-estimator (B) (\tabref{tab:bundlefusion_quantitative} row 10 in BundleFusion) provide significant improvements.
In the presence of heavy occlusions and motion ambiguities, learning the trust region step (C) helps to find better local optima which results in large improvements in our experiments, observed when estimating object motion (\tabref{tab:synthetic_quantitative} row 13) and when estimating camera motion in dynamic scenes (\tabref{tab:bundlefusion_quantitative} row 11 in DynamicBundleFusion).
We find our trust region network (C) to be highly effective for learning accurate relative motion in the presence of large variations in motion magnitude(\tabref{tab:tum_quantitative}).


%





\section{Conclusion}


We have taken a deeper look at the inverse compositional algorithm by rephrasing it as a neural network with three trainable submodules which allows for relaxing some of the most significant restrictions of the original formulation.
Experiments on the challenging task of relative rigid motion estimation from two RGB-D frames demonstrate that our method achieves more accurate results compared to both classical IC algorithms as well as data hungry direct image-to-pose regression techniques.
Although our data-driven IC method can better handle challenges in large object motions, heavy occlusions and varying illuminations, solving those challenges in the wild remains a subject for future research.
To extend the current method to real-world environments with complex entangled motions, possible future directions include exploring multiple motion hypotheses, multi-view constraints and various motion models which will capture a broader family of computer vision tasks.

\clearpage

{\small
	\bibliographystyle{ieee}
	\bibliography{bibliography_custom,bibliography_long,bibliography}
}

\clearpage


\appendix
\section*{Supplementary Materials} 

\noindent In this supplementary document, we first provide details regarding the network architectures used in our model. Next, we show additional experiments on the 2D affine transformation estimation task. Then we provide details on the Jacobian implementation for the rigid motion estimation. We also present more details on our TUM RGB-D experiments as well as the generation of the MovingObjects3D dataset. Finally, we provide additional qualitative experiments for our method on the MovingObjects3D dataset.

\section{Network Architectures}

In the following, we describe the network architectures used in our model.

\begin{figure*}[t!]
	\centering
	\includegraphics[width=\textwidth]{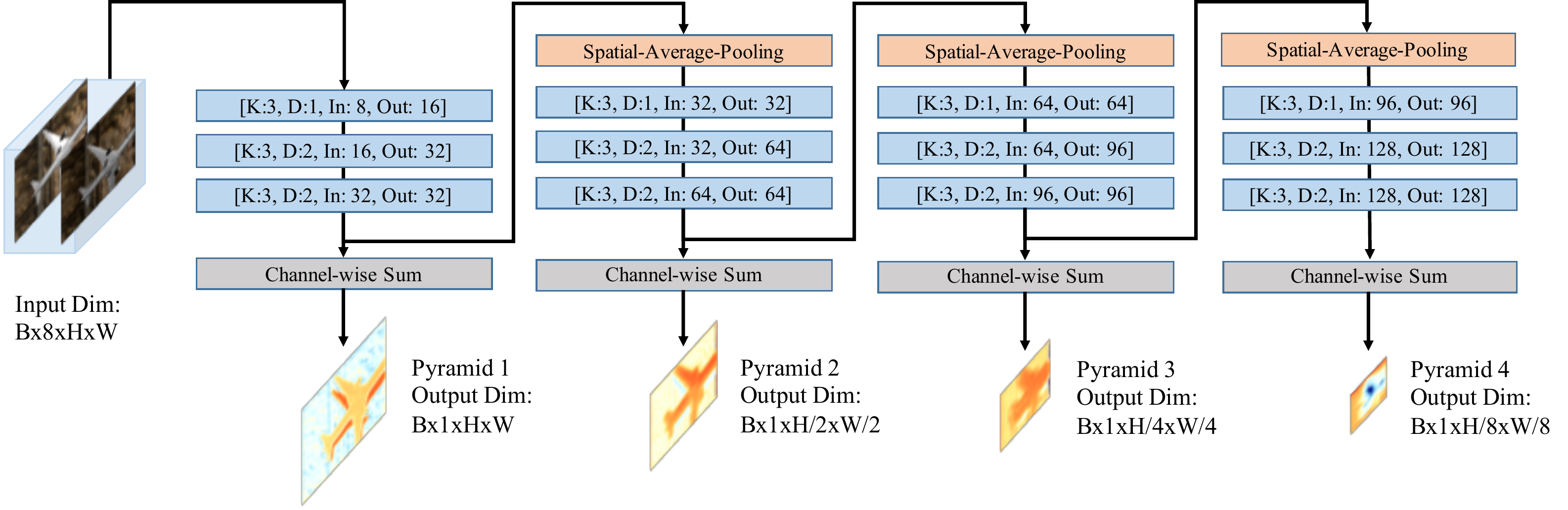}
	\caption{\textbf{(A) Two-view Feature Encoder}. We use [K, D, In, Out] as abbreviation for [Kernel size, Dilation, Input channel size, Output channel size]. All convolutional layers are followed by a BatchNorm layer and a ReLU layer. We use [B,H,W] as abbreviation for the feature size [Batch size, Height of feature, Width of feature]. We use spatial average pooling of size 2 to downsample features between two feature pyramids. We channel-wise sum the output features of the encoder at each scale to obtain the resulting feature maps.}
	\label{fig:two_view_feature_encoder}
\end{figure*}

\boldparagraph{(A) Two-view Feature Encoder} \figref{fig:two_view_feature_encoder} shows the architecture of our two-view feature encoder for estimating both $\bT_{\theta}$ and $\bI_{\theta}$. The network takes two concatenated RGB-D views as input. For the depth channel, we use the inverse depth $d$ clamped to [0, 10].  For the 2D affine experiments we use only the RGB channels.

The feature pyramid comprises one (A) Two-view Feature Encoder at each of the four pyramid levels. Each feature encoder uses three dilated convolutional layers. We use a Spatial Average Pooling layer to downsample the output features from the fine scale as input to the next coarser scale. 

\begin{figure*}[t!]
	\centering
	\includegraphics[width=\textwidth]{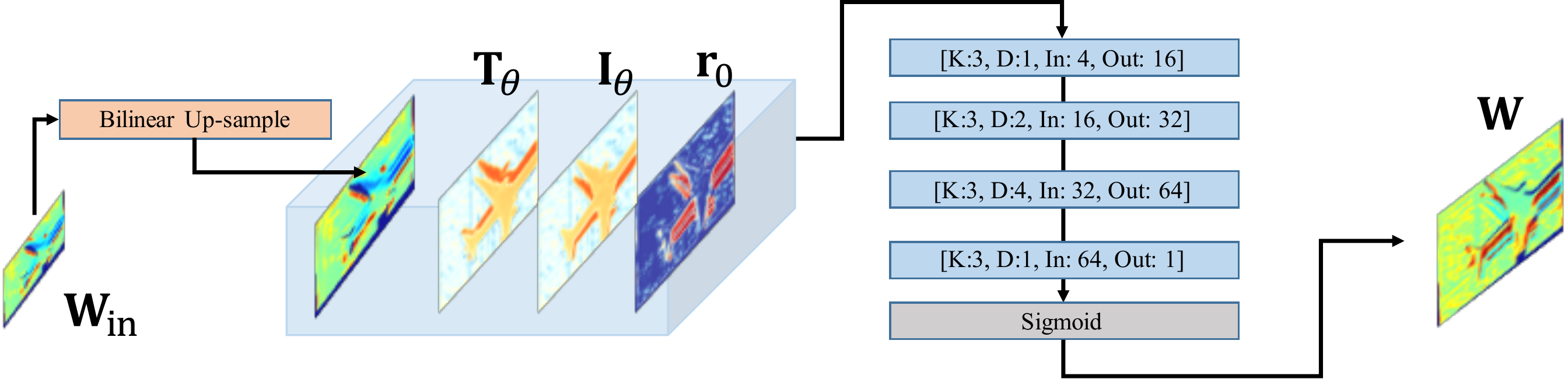}
	\caption{\textbf{(B) Convolutional M-Estimator}. We use [K, D, In, Out] as abbreviation for [Kernel size, Dilation, Input channel size, Output channel size]. All convolutional layers are followed by a BatchNorm layer and a ReLU layer. In our default weight-sharing setting, all weights are shared across networks in different pyramids. At the coarsest image level which does not require the up-sampled $\bW$ as input, we set $\bW_{\text{in}}$ to $\mathbf{1}$.}
	\label{fig:mestimator}
\end{figure*}

\boldparagraph{(B) Convolutional M-Estimator} \figref{fig:mestimator} shows the operations and parameters of the Convolutional M-Estimator we use in this paper. In the coarse-to-fine inverse compositional refinement, we add one more input to the network which is the predicted weight from the coarser level pyramid. At each image pyramid, we bilinear upsample the weight matrix predicted from the coarse scale $\bW_{\textbf{in}}$, and concatenate it with $\bT_{\theta}$, $\bI_{\theta}$ and $\br_{0}$, which we use as input to the Convolutional M-estimator. The network predicts $\bW$ at the current scale. Different from traditional M-estimators which evaluate $\bW$ at every step when $\br_{k}$ is updated, we only compute $\bW$ once for all following $K$ iterations. This way, we approximate the classical M-estimator and significantly reduce computation. 
The network is composed of four convolutional layers, with dilation [1,2,4,1], followed by a sigmoid layer which normalizes the output to the range [0,1].
Note that despite the small size of our network, the dilation layers and the coarse-to-fine process ensure a sufficiently large receptive field.

\begin{figure*}[t!]
	\centering
	\includegraphics[width=0.90\textwidth]{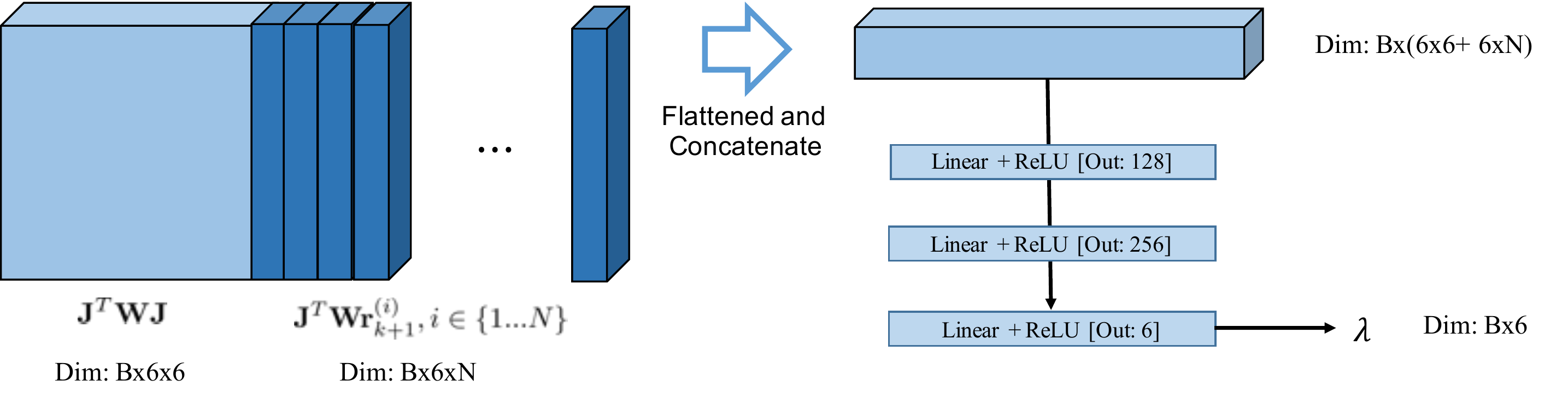}
	\caption{\textbf{(C) Trust Region Network. } We use B as abbreviation for Batch size. $N$ indicates the number of damping proposals. In the weight-sharing setting, all weights are shared across networks at different pyramid levels. We use 'Linear' to represent a fully connected layer. The last ReLU layer ensures that the output $\lambda$ is non-negative.}
	\vspace{0.5cm}
	\label{fig:trust_region_network}
\end{figure*}

\boldparagraph{(C) Trust Region Network} \figref{fig:trust_region_network} shows the operations and parameters of our Trust Region Network. Given the $N$ residual maps $\br_{(i)}^k, i \in \{1...N\}$, we first calculate the right-hand-side (RHS) vector $\bJ^{T}\bW \br^{(i)}_k \in \nR^{1\times6}$ corresponding to each residual map $\br^{(i)}_k$. Next, we flatten the $N$ RHS vectors jointly with the approximate Hessian matrix $\bJ^{T}\bW \bJ \in \nR^{6\times6}$ into a single vector, which is the input to our Trust Region Network. This network is composed of three fully connected layers and outputs the damping vector. At the last layer, a ReLU ensures non-negative elements.

\section{2D Affine Motion Estimation}

The proposed framework is general and can be applied to a wide range of motion models apart from the 3D rigid motion estimation tasks presented in the main paper, see, \eg, \cite{Szeliski07ftcgv}.
To demonstrate its generality, this section provides results on 2D affine motion estimation.
While 2D affine motion estimation is in general easier than 3D rigid motion estimation, occlusions cannot be treated explicitly as depth is unknown. Thus, any successful method must implicitly identify occlusions as outliers during estimation. 

\boldparagraph{Implementation} Given pixel $\bx=(x,y)^{T} \in \nR^{2}$, we define the warping function $\bW_{\bxi}$ using the following parameterization (see also \cite{Baker04ijcv})
\begin{equation}
\begin{bmatrix}  1+\xi_{1} & \xi_{3} & \xi_{5} \\
\xi_{2} & 1+\xi_{4} & \xi_{6} 
\label{eq:affine}
\end{bmatrix} 
\end{equation}
where $\bxi = (\xi_{1}, \xi_{2}, \xi_{3}, \xi_{4}, \xi_{5}, \xi_{6})^{T}$. The analytic form of the Jacobian in \eqref{eq:affine} is 
\begin{equation}
\begin{bmatrix}
x & 0 & y & 0 & 1 & 0 \\
0 & x & 0 & y & 0 & 1
\end{bmatrix}
\end{equation}
The composition of $\bxi \circ \Delta \bxi$ is given by
\begin{equation}
\begin{bmatrix}
\xi_{1} + \Delta \xi_{1} + \xi_{1} \Delta \xi_{1} + \xi_{3}\Delta \xi_{2} \\
\xi_{2} + \Delta \xi_{2} + \xi_{2} \Delta \xi_{1} + \xi_{4}\Delta \xi_{2} \\
\xi_{3} + \Delta \xi_{3} + \xi_{1} \Delta \xi_{3} + \xi_{3}\Delta \xi_{4} \\
\xi_{4} + \Delta \xi_{4} + \xi_{2} \Delta \xi_{3} + \xi_{4}\Delta \xi_{4} \\
\xi_{5} + \Delta \xi_{5} + \xi_{1} \Delta \xi_{5} + \xi_{3}\Delta \xi_{6} \\
\xi_{6} + \Delta \xi_{6} + \xi_{2} \Delta \xi_{5} + \xi_{4}\Delta \xi_{6} \\
\end{bmatrix}
\end{equation}
The parameters of the inverse affine $\xi^{-1}$ in \eqref{eq:affine} are 
\begin{equation}
	\frac{1}{(1+\xi_1)(1+\xi_4) - \xi_2\xi_3} \begin{bmatrix}
	-\xi_{1} - \xi_{1}\xi_{4} + \xi_{2}\xi_{3} \\
	-\xi_{2} \\ 
	-\xi_{3} \\
	-\xi_{4} - \xi_{1}\xi_{4} + \xi_{2}\xi_{3} \\
	-\xi_{5} - \xi_{4}\xi_{5} + \xi_{3}\xi_{6} \\
	-\xi_{6} - \xi_{1}\xi_{6} + \xi_{2}\xi_{5}
	\end{bmatrix}
\end{equation}

\begin{figure}[b]
\centering
\begin{tabularx}{\columnwidth}{cccc}
$\bT$ & 
\parbox[c]{0.25\columnwidth}{\includegraphics[width=0.28\columnwidth]{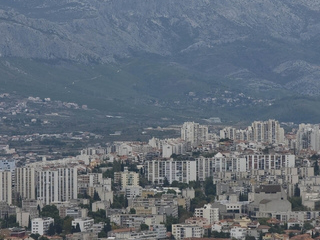}}&
\parbox[c]{0.25\columnwidth}{\includegraphics[width=0.28\columnwidth]{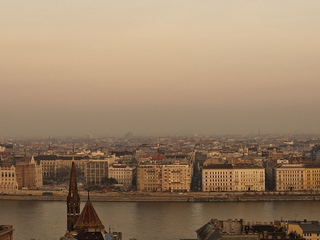}}&
\parbox[c]{0.25\columnwidth}{\includegraphics[width=0.28\columnwidth]{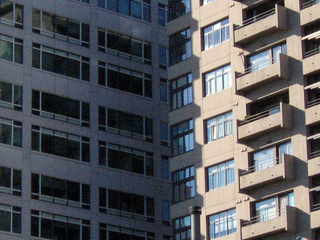}} \\
\\
$\bI$ & 
\parbox[c]{0.25\columnwidth}{\includegraphics[width=0.28\columnwidth]{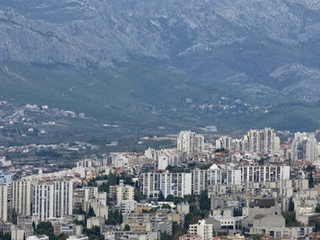}}&
\parbox[c]{0.25\columnwidth}{\includegraphics[width=0.28\columnwidth]{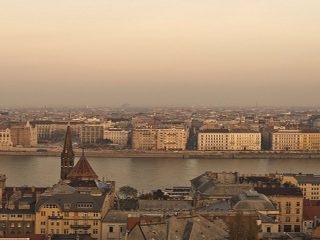}}&
\parbox[c]{0.25\columnwidth}{\includegraphics[width=0.28\columnwidth]{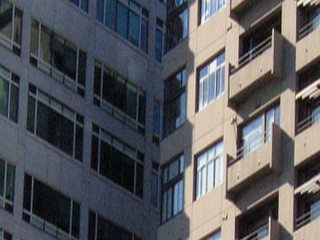}}
\end{tabularx}
\caption{Examples of $\bT$ and $\bI$ used for our 2D affine transformation estimation experiments.}
\label{fig:affine_examples}
\end{figure}

\boldparagraph{Datasets} We download 1800 natural images from flickr and use them to generate a 2D dataset for training affine 2D transformation estimation. We use the downloaded flickr images as $\bT$. Given a random affine transform which induces a 2D warping field, we synthesize the template $\bI$ by applying the warping field to $\bT$ using bilinear interpolation. To remove the boundary effects caused by zero padding in the warping process, we crop a region of size 240x320 from the central region of both $\bI$ and $\bT$. According to the actual raw image size and the coordinates of the cropped region, we adjust the warping field to ensure the cropped region has the correct affine transform. \figref{fig:affine_examples} shows examples of generated pairs $\bT$ and $\bI$.

For each template image, we randomly generate six different affine transforms, and synthesize six different images $\bI$ using the warping field induced from each affine transform. The generated affine transforms are used as ground truth for training and evaluation. We use five out of the six generated affine transforms and their corresponding image pairs as the training set, and use the rest for testing. 

\boldparagraph{Baselines} We follow the experiment settings in the main paper. We use (A), (B), (C) to refer to our contributions in Sec.1. We set $\bW$ to the identity matrix when the Convolutional M-Estimator (B) is not used and use Gauss-Newton optimization in the absence of the Trust Region
Network (C). We consider the following configurations: 
\begin{itemize}[noitemsep,nolistsep,leftmargin=*]
\item \textbf{Ours (A)+(B)+(C):} Our proposed method with shared
weights. We perform coarse-to-fine iterations on three
pyramid levels with three IC iterations at each level. We
use shared weights for all iterations in (B) and (C). 
\item \textbf{DeepLK-6DoF} We implemented a variant of DeepLK
\cite{Wang18icra} which predicts the affine transform instead
of translation and scale as in the original 2D task. We use Gauss-Newton as the default optimization
for this approach and no Convolutional M-Estimator.
Comparing this approach with our method when
using only the two-view feature network (A) demonstrates the
utility of our two-view feature encoder.
\end{itemize}

\boldparagraph{Training} During training, we minimizes the L1 norm of the distance from our estimated affine transform $\bxi^{\star}$ to the ground truth affine transform $\bxi^{\text{GT}}$. We train our method and all baselines using a learning rate of 0.005. 

\begin{table}[t]
	\centering
	\begin{tabular}{@{}lc@{}}
		\toprule
		& L1 error \\ \midrule
		No Learning          &   0.219       \\
		DeepLK, adapted from \cite{Wang18icra} &   0.158       \\
		Ours: (A)            &   0.151       \\
		Ours: (A)+(B)        &   0.132       \\
		Ours: (A)+(B)+(C)    &   0.071   \\ \bottomrule
	\end{tabular}
\caption{Quantitative evaluation on the test set using L1 error wrt. the estimated 2D affine transform.}
\label{tab:affine_results}
\end{table}

\boldparagraph{Results} \tabref{tab:affine_results} shows a quantitative evaluation of our 2D affine motion estimation experiments. We evaluate the error in L1 norm by comparing the estimated results to the ground truth. Compared to all baseline methods, our model ((A)+(B)+(C)) yields the most accurate solution. 

Note that different from the 3D motion estimation experiments in the main paper, there exists no motion ambiguity in the datasets used for 2D affine transform, rendering this setting simpler. We observe small improvements when using our two-view feature encoder (A) compared to using only a single view \cite{Wang18icra}. Using (B) Convolutional M-estimator gives better performance which may potentially address outliers induced by occlusions. Using (C) Trust Region Network further helps the network to boost performance. Similarly to our results in the main paper, we observe that we obtain the most accurate results when combining all components ((A)+(B)+(C)) of our model.

\section{Implementation Details of Jacobian}

Given pixel $\bx \in \nR^{2}$, camera intrinsic $\bK$ and depth $D(\bx)$, we define the corresponding 3D point $\bp=(p_x, p_y, p_z)$ as $\bp=D(\bx)\bK^{-1}\bx$. At each iteration, we apply an exponential map to the output of the network $\xi \in \se{3}$ to obtain the transformation matrix $\bT_{\xi}=\exp([\xi]_{\times})$. Suppose $\bK$ is the intrinsic matrix for a pin-hole camera without distortion, which can be parameterized as $[f_x, f_y, c_x, c_y]$ with $f_x, f_y$ as its focal length and $c_x, c_y$ as its offset along the two axes. The Jacobian $\bJ$  of template image $\bT(\mathbf{0})$ with respect to warp parameter $\xi$ is given by
\begin{align}
	\bJ &= \nabla \bT \frac{\partial \bW}{\partial \bxi}  \label{eq:Jacobian_overall} \\
	\nabla \bT &= \begin{bmatrix}
	\frac{\partial \bT(\mathbf{0})}{\partial u} & \frac{\partial \bT(\mathbf{0})}{\partial v}
	\end{bmatrix} \\
	\frac{\partial \bW}{\partial \bxi} &= \begin{bmatrix}
	f_x \\ f_y
	\end{bmatrix} \cdot
	\begin{bmatrix}
	-\frac{p_x p_y}{p_z^2} & 1+\frac{p_x^2}{p_y^2} & -\frac{p_y}{p_z} & \frac{1}{p_z} & 0 & -\frac{p_x}{p_z^2} \\
	-1 - \frac{p_y^2}{p_z^2} & - \frac{p_x p_y}{p_z^2} & \frac{p_x}{p_z} & 0 & \frac{1}{p_z} & -\frac{p_y}{p_z^2}
	\end{bmatrix}
	\label{eq:warping_jacobian}
\end{align}
where $\cdot$ is the element-wise product along each row. $\nabla \bT$ is the image gradient of template $\bT(\mathbf{0})$. In the proposed model, we compute the Jacobian of $\bT_{\theta}$ from the output of network (A) by substituting $\bT$ with $\bT_{\theta}$ in \eqref{eq:Jacobian_overall}.

We further simplify \eqref{eq:warping_jacobian} by exploiting the inverse depth parameterization $\bp=(p_u / p_d, p_v / p_d, 1/p_d)$ where $(p_u, p_v) \in \nR^2$ is the pixel coordinate and $p_d \in \nR^1$ is the inverse depth.
We obtain:
%
\begin{equation}
	\frac{\partial \bW}{\partial \bxi} =  \begin{bmatrix}
		f_x \\ f_y
	\end{bmatrix} \cdot
	\begin{bmatrix}
		-p_u p_v  &  1+p_u^2 & -p_v & p_d & 0 & -p_d p_u \\
		-1-p_v^2  &  p_u p_v  &  p_u  & 0 	 & p_d & -p_d p_v
	\end{bmatrix} 
\end{equation}
For more details we refer the reader to \cite{Blanco10se3}.

\section{Details of TUM RGB-D Experiments}

Our evaluation split of the TUM RGB-D SLAM dataset \cite{Sturm12iros} consists of four trajectories of different conditions, trajectory length and motion magnitudes. After synchronization of the color image, depth and the ground truth trajectory, we obtain 750 frames in 'fr1/360', 584 frames in 'fr1/desk', 2203 frames in 'fr2/desk' and 830 frames in 'fr2/pioneer\_360'. To better understand our method operating under different conditions, we present a detailed quantitative evaluation in  \tabref{tab:tum_quantitative_per_trajectory} using the Relative Pose Error (RPE) and the 3D End Point Error (3D EPE) metrics for each trajectory. 

\boldparagraph{Discussion} Our method with all modules (A)+(B)+(C) outperforms the baseline RGBD visual odometry \cite{Steinbrucker11iccvw} across all subsampled trajectories, except in 'fr1/desk' of keyframe [1,2] and 'fr2/desk' of keyframe 2 in which the performance is close to each other. Our method shows clear advantages when the motion magnitude is large, e.g. 'fr1/360', 'fr2/pioneer\_360'. From the ablation, we also observe that adding the trust-region module (C) helps to stabilize training and yields the most significant improvement in test accuracy. One possible reason for this is that the trust-region module can adjust the damping parameters and adapt to a wide range of motion magnitudes in the data. At training time, adaptive damping helps to stabilize the training loss and potentially contributes to learning better features in modules (A) and (B). At inference time, the method can adjust its trust-region step to adapt to different motion magnitudes with a fixed number of iterations. 



\begin{table*}[t]
	\centering
	\small
	\begin{tabular}{@{}llcccccccc@{}}
		\toprule
		& \multicolumn{1}{c}{\multirow{2}{*}{}} 
		& \multicolumn{4}{c}{\begin{tabular}[c]{@{}c@{}} mRPE: $\theta$ (Deg) $\downarrow$ / $t$ (cm) $\downarrow$  \end{tabular} } & \multicolumn{4}{c}{3D EPE (cm) }   \\				
		& & KF 1  & KF 2 &  KF 4 & KF 8 &  KF 1  & KF 2 &  KF 4 & KF 8  \\ \midrule

		& \multicolumn{9}{c}{fr1/360}    \\ \midrule
		& RGBD VO \cite{Steinbrucker11iccvw} & 0.46/1.03 & 2.45/5.26 & 7.47/10.31 & 16.08/17.32 & 1.33 & 5.98 & 21.34 & 52.50  \\
		& Ours: (A) & 0.50/1.33 & 1.32/3.84  & 5.68/11.79 & 14.33/16.26 & 1.24 & 2.39 & 13.37 & 49.60  \\
		& Ours: (A)+(B) & 0.45/1.18 & 1.00/3.18 & 4.96/11.62 & 14.23/17.52 & 1.18 & 1.92 & 10.67 & 49.13  \\
		& Ours: (A)+(B)+(C) & \textbf{0.33}/\textbf{0.61}  & \textbf{0.49}/\textbf{1.20} & \textbf{2.64}/\textbf{2.63} & \textbf{7.24}/\textbf{6.64} & \textbf{1.05} & \textbf{1.21} & \textbf{2.64} & \textbf{22.40}  \\ \midrule

		& \multicolumn{9}{c}{fr1/desk}    \\ \midrule
		& RGBD VO \cite{Steinbrucker11iccvw} & \textbf{0.43}/\textbf{0.69} &  \textbf{0.76}/\textbf{1.04} & 3.52/5.15 & 10.71/19.83 & \textbf{0.59} & \textbf{0.91} & 5.46 & 18.84 \\
		& Ours: (A) & 0.58/1.04 & 1.12/2.05 & 2.75/4.87 & 7.14/11.27 & 0.74 & 1.31 & 3.89 & 12.89 \\
		& Ours: (A)+(B) & 0.57/1.02 & 1.08/2.03  & 2.54/4.63 & 6.59/10.87 & 0.74 & 1.28 & 3.42 & 11.21 \\
		& Ours: (A)+(B)+(C) & 0.55/0.90  & 0.89/1.55 & \textbf{1.58}/\textbf{2.59} & \textbf{4.30}/\textbf{7.11} & 0.68 & 0.96 & \textbf{1.74} & \textbf{6.11} \\ \midrule
		
		& \multicolumn{9}{c}{fr2/desk}    \\ \midrule
		& RGBD VO \cite{Steinbrucker11iccvw} &  0.30/0.56 & \textbf{0.35}/\textbf{0.72} & 0.79/1.78 & 1.44/3.80 & 0.96 & \textbf{0.93} & 2.20 & 4.75 \\
		& Ours: (A) & 0.30/0.54 & 0.44/0.98 & 0.72/1.94 & 1.46/4.30 & 0.80 & 1.04 & 1.60 & 2.93 \\
		& Ours: (A)+(B) & 0.31/0.56 & 0.45/1.03 & 0.77/2.04 & 1.43/4.14 & 0.80 & 1.04 & 1.62 & 2.83  \\
		& Ours: (A)+(B)+(C) & \textbf{0.27}/\textbf{0.42}  & 0.39/0.73 & \textbf{0.61}/\textbf{1.37} & \textbf{1.11}/\textbf{2.79} & \textbf{0.73} & 0.94 & \textbf{1.29} & \textbf{2.09} \\ \midrule
				
		& \multicolumn{9}{c}{fr2/pioneer\_360}    \\ \midrule
		& RGBD VO \cite{Steinbrucker11iccvw} &  1.02/1.82 & 2.00/4.19 & 4.20/6.51 & 8.54/14.36 & 6.39 & 9.22 & 21.16 & 48.46 \\
		& Ours: (A) & 0.76/1.77 & 1.04/3.66 & 2.34/8.16 & 7.60/14.83 & 3.55 & 5.67 & 13.78 & 39.41 \\
		& Ours: (A)+(B) & 0.72/1.80 & 0.95/3.52 & 2.15/6.96 & 6.91/13.31 & 3.57 & 5.42 & 12.77 & 36.80 \\
		& Ours: (A)+(B)+(C) & \textbf{0.65}/\textbf{0.85}  & \textbf{0.74}/\textbf{1.07} & \textbf{0.98}/\textbf{1.79} & \textbf{2.38}/\textbf{6.85} & \textbf{2.76} & \textbf{3.15} & \textbf{4.46} & \textbf{13.52} \\ \midrule
				
		& \multicolumn{9}{c}{Average over trajectories mRPE and EPE}    \\ \midrule
		& RGBD VO \cite{Steinbrucker11iccvw} &  0.55/1.03 & 1.39/2.81 & 3.99/5.95 & 9.20/13.83  & 2.31 & 4.38 & 12.67 & 31.13   \\
		& Ours: (A) & 0.53/1.17 & 0.97/2.63 & 2.87/6.89 & 7.63/12.16& 1.58 & 2.60 & 8.15 & 26.20 \\
		& Ours: (A)+(B) & 0.51/1.14 & 0.87/2.44 & 2.60/6.56 & 7.30/11.21 & 1.56 & 2.41 & 7.10 & 24.69 \\
		& Ours: (A)+(B)+(C) & \textbf{0.45}/\textbf{0.69}  & \textbf{0.63}/\textbf{1.14} & \textbf{1.10}/\textbf{2.09} & \textbf{3.76}/\textbf{5.88} & \textbf{1.31} & \textbf{1.57} & \textbf{2.53} & \textbf{11.03} \\ \midrule
		
		& \multicolumn{9}{c}{Average over frames mRPE and EPE}    \\ \midrule
		& RGBD VO \cite{Steinbrucker11iccvw} & 0.48/0.90 & 1.08/2.20 & 2.95/4.60 & 6.54/10.26 & 1.80 & 3.53 & 9.58 & 23.14  \\
		& Ours: (A) & 0.46/0.98 & 0.79/2.12 & 2.16/5.35 & 5.58/9.65 & 1.39 & 2.18 & 6.22 & 19.16 \\
		& Ours: (A)+(B) & 0.45/0.96 & 0.73/2.00 & 1.99/5.15 & 5.35/9.41 & 1.38 & 2.05 & 5.52 & 18.31  \\
		& Ours: (A)+(B)+(C) & \textbf{0.39}/\textbf{0.59}  & \textbf{0.54}/\textbf{0.98} & \textbf{0.91}/\textbf{1.83} & \textbf{2.82}/\textbf{4.80} & \textbf{1.16} & \textbf{1.41} & \textbf{2.18} & \textbf{8.28} \\
		
		\bottomrule
	\end{tabular}
	\caption{\textbf{Detailed Results on TUM RGB-D Dataset [45].} This table shows the mean relative pose error (mRPE) on our test split of the TUM RGB-D Dataset [45]. KF denotes the size of the key frame intervals.}
	\label{tab:tum_quantitative_per_trajectory}
\end{table*}

\section{Details of MovingObjects3D}


\begin{figure}[t]
	\subfloat[][aeroplane] {\includegraphics[width=0.33\columnwidth]{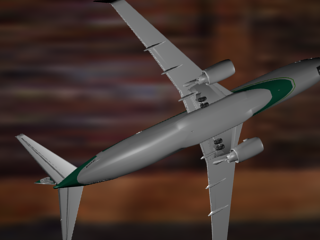} }
	\subfloat[][boat] {\includegraphics[width=0.33\columnwidth]{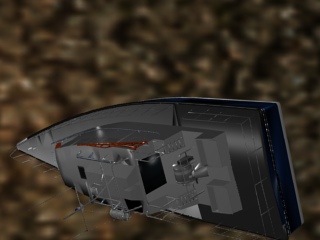} } 
	\subfloat[][bus] {\includegraphics[width=0.33\columnwidth]{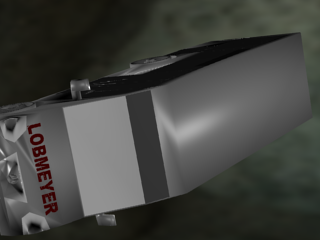} } \\
	\subfloat[][car] {\includegraphics[width=0.33\columnwidth]{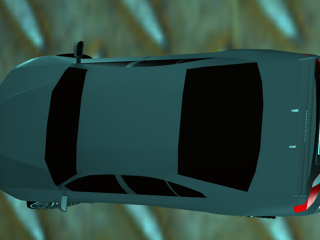} }
	\subfloat[][motorbike] {\includegraphics[width=0.33\columnwidth]{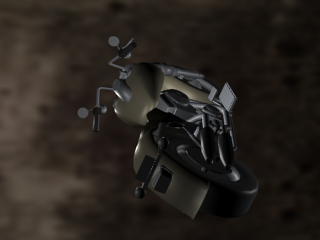} } 
	\subfloat[][bicycle] {\includegraphics[width=0.33\columnwidth]{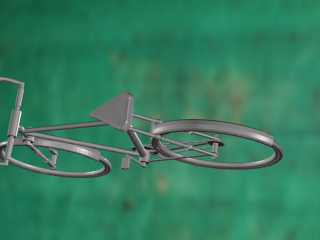} }
	\caption{Example images from the MovingObjects3D dataset. We use 'boat' and 'motorbike' categories as test set and the others as train/validation set.}
	\label{fig:example_MovingObjs3D}
\end{figure}

\boldparagraph{Background Set-up} We use a cubic box of size $6\times6\times6$ as the background layout. We download 150 background textures from TextureNinja\footnote{https://texture.ninja/} and use them as the textures for the inner side of the cube. We randomly locate the camera inside the cube with zero elevation. We randomly place four point light sources at four different directions. 

\boldparagraph{Object Trajectories} For each trajectory, we randomly generate keyframe poses (we use one keyframe every ten frames). We calculate poses between keyframes using the default trajectory interpolation in Blender. We randomly locate the objects within a 1.5m radius ball around the center of the cube which ensures that the objects always move inside of the box. There is a significant number of views for which only parts of the object is visible. We exclude all frames where the entire object is outside the field of view of the camera.

\boldparagraph{Rendered Images and Ground Truth} We render the images, depth maps and instance segmentation masks using Blender. We show examples of objects for all six categories in \figref{fig:example_MovingObjs3D}. We use images rendered using the categories ’boat’ and ’motorbike’ as test set and images from the categories ’aeroplane’, ’bicycle’, ’bus’, ’car’ as the training set. We will make the dataset creation tool and the dataset itself public upon publication.

\section{Qualitative Visualizations}

\begin{figure*}[t]
	\centering 
	\begin{tabularx}{\textwidth}{ccccc}
	\vspace{0.05cm}
	$\bT$ & 
	\parbox[c]{0.20\textwidth}{\includegraphics[width=0.22\textwidth]{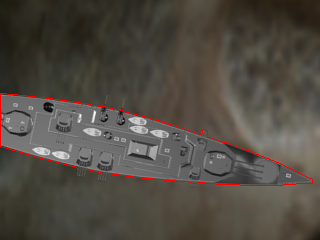}}&
	\parbox[c]{0.20\textwidth}{\includegraphics[width=0.22\textwidth]{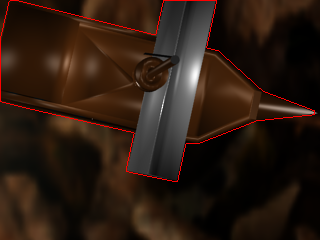}}&
	\parbox[c]{0.20\textwidth}{\includegraphics[width=0.22\textwidth]{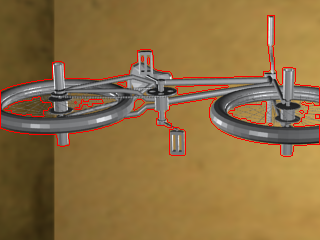}}&
	\parbox[c]{0.20\textwidth}{\includegraphics[width=0.22\textwidth]{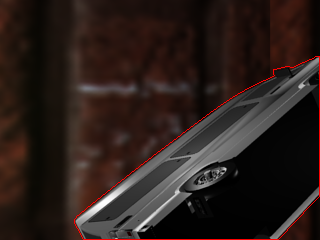}}
	\\ \vspace{0.05cm}
	$\bI$ & 
	\parbox[c]{0.20\textwidth}{\includegraphics[width=0.22\textwidth]{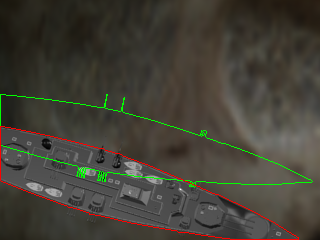}}&
	\parbox[c]{0.20\textwidth}{\includegraphics[width=0.22\textwidth]{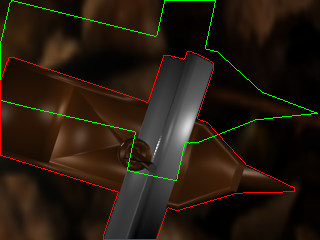}}&
	\parbox[c]{0.20\textwidth}{\includegraphics[width=0.22\textwidth]{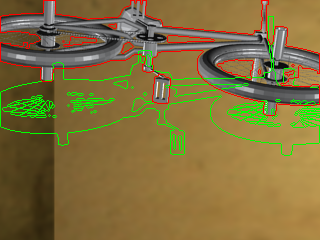}}&
	\parbox[c]{0.20\textwidth}{\includegraphics[width=0.22\textwidth]{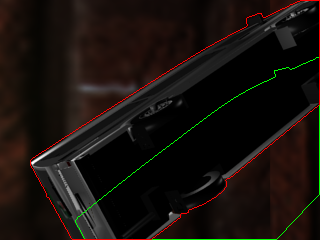}}
	\\ \vspace{0.05cm}
	$\bI(\bxi^{\star}_{1})$ & 
	\parbox[c]{0.20\textwidth}{\includegraphics[width=0.22\textwidth]{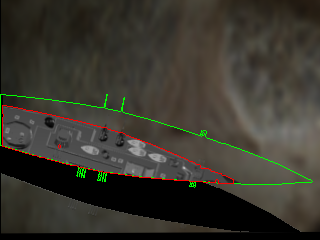}}&
	\parbox[c]{0.20\textwidth}{\includegraphics[width=0.22\textwidth]{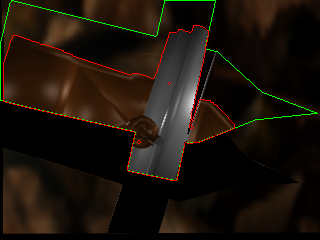}}&
	\parbox[c]{0.20\textwidth}{\includegraphics[width=0.22\textwidth]{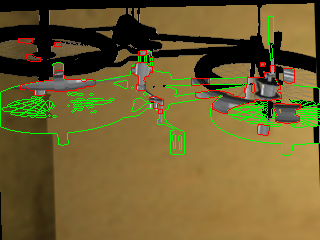}}&
	\parbox[c]{0.20\textwidth}{\includegraphics[width=0.22\textwidth]{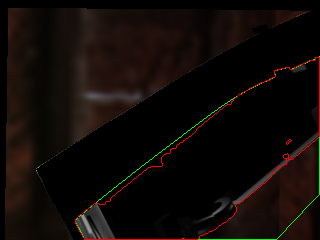}}
	\\ \vspace{0.05cm}
	$\bI(\bxi^{\star}_{2})$ & 
	\parbox[c]{0.20\textwidth}{\includegraphics[width=0.22\textwidth]{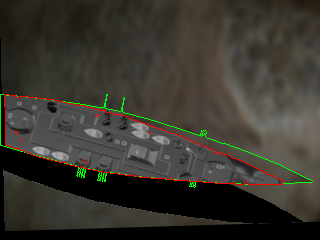}}&
	\parbox[c]{0.20\textwidth}{\includegraphics[width=0.22\textwidth]{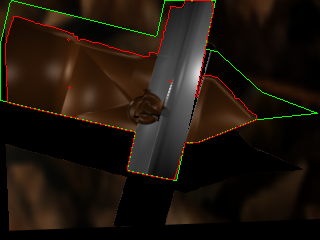}}&
	\parbox[c]{0.20\textwidth}{\includegraphics[width=0.22\textwidth]{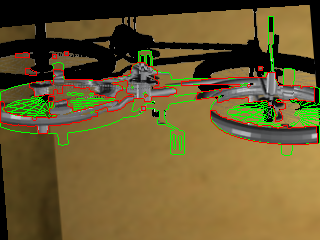}}&
	\parbox[c]{0.20\textwidth}{\includegraphics[width=0.22\textwidth]{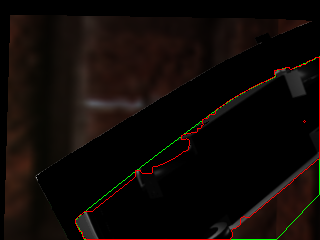}}
	\\ \vspace{0.05cm}
	$\bI(\bxi^{\star}_{3})$ & 
	\parbox[c]{0.20\textwidth}{\includegraphics[width=0.22\textwidth]{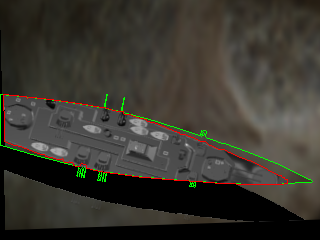}}&
	\parbox[c]{0.20\textwidth}{\includegraphics[width=0.22\textwidth]{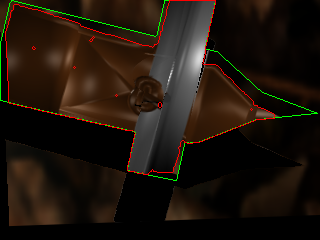}}&
	\parbox[c]{0.20\textwidth}{\includegraphics[width=0.22\textwidth]{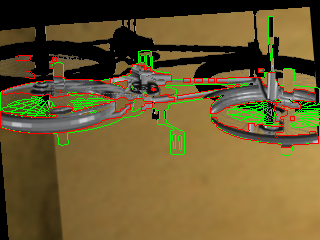}}&
	\parbox[c]{0.20\textwidth}{\includegraphics[width=0.22\textwidth]{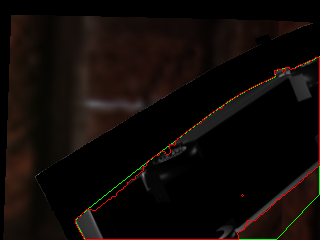}}
	\\ \vspace{0.05cm}
	$\bI(\bxi^{\star}_{\text{final}})$ & 
	\parbox[c]{0.20\textwidth}{\includegraphics[width=0.22\textwidth]{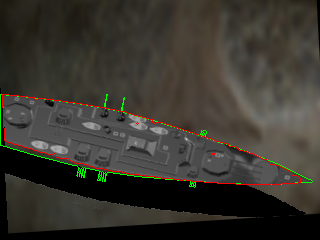}}&
	\parbox[c]{0.20\textwidth}{\includegraphics[width=0.22\textwidth]{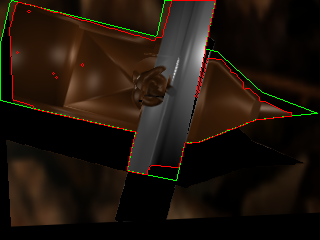}}&
	\parbox[c]{0.20\textwidth}{\includegraphics[width=0.22\textwidth]{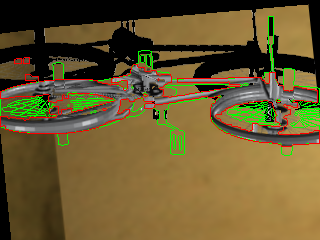}}&
	\parbox[c]{0.20\textwidth}{\includegraphics[width=0.22\textwidth]{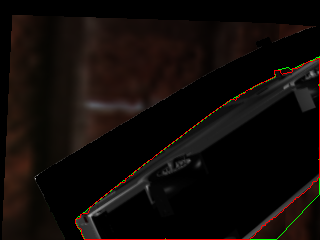}}
	\\ \vspace{0.05cm}
	$\bI(\bxi^{\text{GT}})$ & 
	\parbox[c]{0.20\textwidth}{\includegraphics[width=0.22\textwidth]{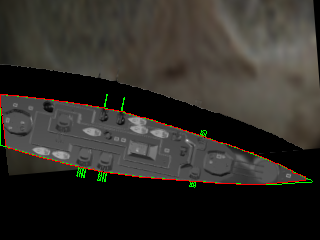}}&
	\parbox[c]{0.20\textwidth}{\includegraphics[width=0.22\textwidth]{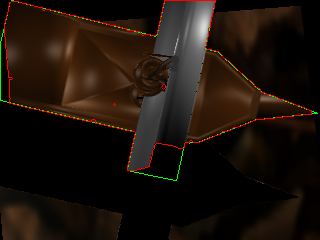}}&
	\parbox[c]{0.20\textwidth}{\includegraphics[width=0.22\textwidth]{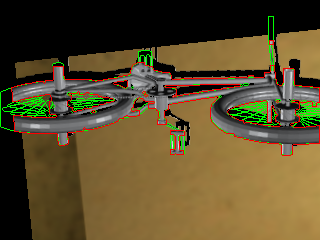}}&
	\parbox[c]{0.20\textwidth}{\includegraphics[width=0.22\textwidth]{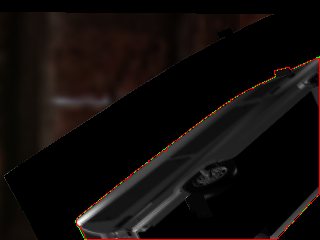}}
	\\ 	
	\end{tabularx}
	\caption{\textbf{Qualitative Results on MovingObjects3D.} Visualization of the warped image $\bI(\bxi)$ using the ground truth object motion $\bxi^{\text{GT}}$ (last row) and the object motion $\bxi^{\star}$ estimated using our method at each pyramid ($\bxi^{\star}_{1}$, $\bxi^{\star}_{2}$, $\bxi^{\star}_{3}$, $\bxi^{\star}_{\text{final}}$) on the MovingObjects3D \emph{validation} and \emph{test} set. In $\bI$, we plot the instance boundary in \textcolor{red}{red} and that of $\bT$ in \textcolor{darkgreen}{green} as comparison. Note the difficulty of the task (truncation, independent background object) and the high quality of our alignments. Black regions in the warped image are due to truncation or occlusion.}
	\label{fig:iterative_vis1}
\end{figure*}

\begin{figure*}[t]
	\centering 
	\begin{tabularx}{\textwidth}{ccccc}
		\vspace{0.05cm}
		$\bT$ & 
		\parbox[c]{0.20\textwidth}{\includegraphics[width=0.22\textwidth]{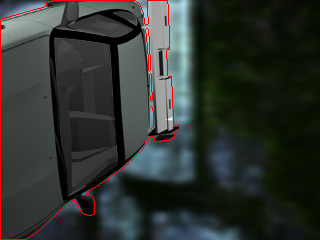}}&
		\parbox[c]{0.20\textwidth}{\includegraphics[width=0.22\textwidth]{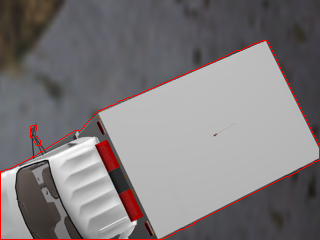}}&
		\parbox[c]{0.20\textwidth}{\includegraphics[width=0.22\textwidth]{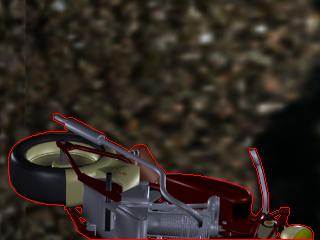}}&
		\parbox[c]{0.20\textwidth}{\includegraphics[width=0.22\textwidth]{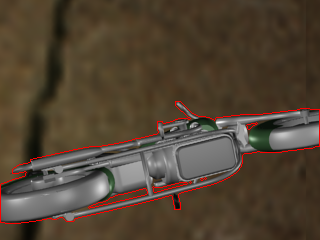}}
		\\ \vspace{0.05cm}
		$\bI$ & 
		\parbox[c]{0.20\textwidth}{\includegraphics[width=0.22\textwidth]{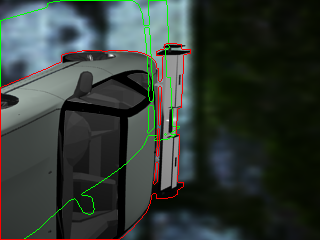}}&
		\parbox[c]{0.20\textwidth}{\includegraphics[width=0.22\textwidth]{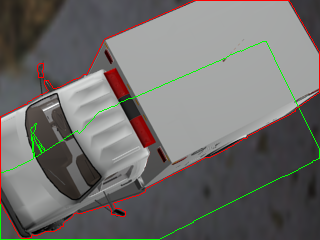}}&
		\parbox[c]{0.20\textwidth}{\includegraphics[width=0.22\textwidth]{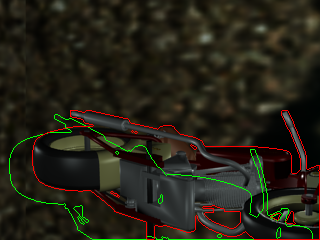}}&
		\parbox[c]{0.20\textwidth}{\includegraphics[width=0.22\textwidth]{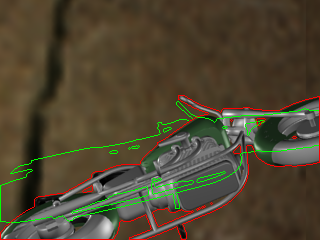}}
		\\ \vspace{0.05cm}
		$\bI(\bxi^{\star}_{1})$ & 
		\parbox[c]{0.20\textwidth}{\includegraphics[width=0.22\textwidth]{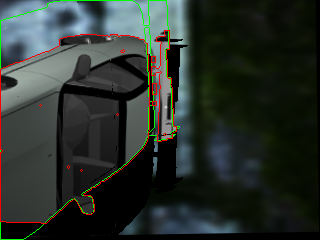}}&
		\parbox[c]{0.20\textwidth}{\includegraphics[width=0.22\textwidth]{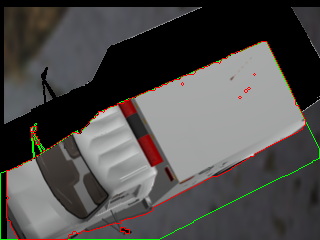}}&
		\parbox[c]{0.20\textwidth}{\includegraphics[width=0.22\textwidth]{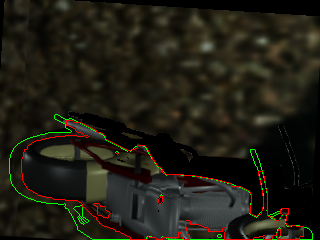}}&
		\parbox[c]{0.20\textwidth}{\includegraphics[width=0.22\textwidth]{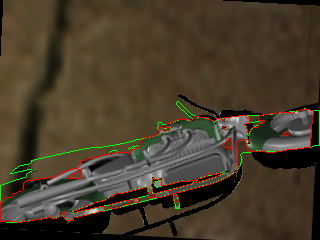}}
		\\ \vspace{0.05cm}
		$\bI(\bxi^{\star}_{2})$ & 
		\parbox[c]{0.20\textwidth}{\includegraphics[width=0.22\textwidth]{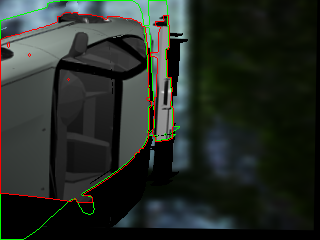}}&
		\parbox[c]{0.20\textwidth}{\includegraphics[width=0.22\textwidth]{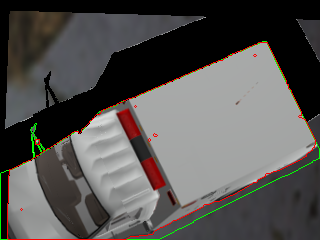}}&
		\parbox[c]{0.20\textwidth}{\includegraphics[width=0.22\textwidth]{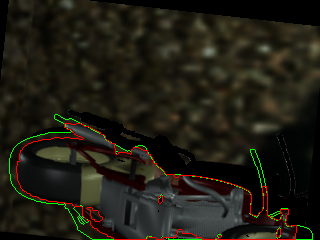}}&
		\parbox[c]{0.20\textwidth}{\includegraphics[width=0.22\textwidth]{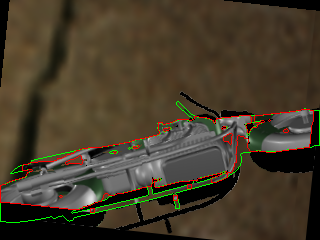}}
		\\ \vspace{0.05cm}
		$\bI(\bxi^{\star}_{3})$ & 
		\parbox[c]{0.20\textwidth}{\includegraphics[width=0.22\textwidth]{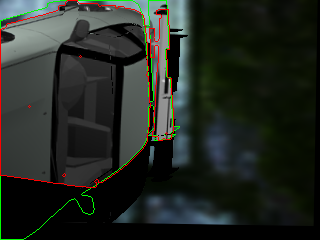}}&
		\parbox[c]{0.20\textwidth}{\includegraphics[width=0.22\textwidth]{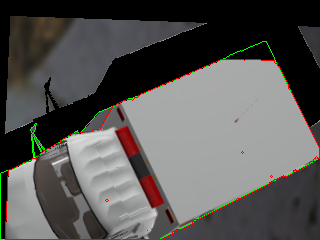}}&
		\parbox[c]{0.20\textwidth}{\includegraphics[width=0.22\textwidth]{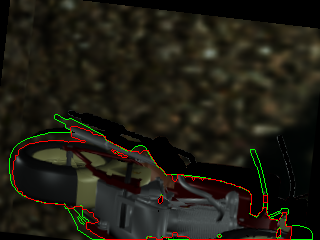}}&
		\parbox[c]{0.20\textwidth}{\includegraphics[width=0.22\textwidth]{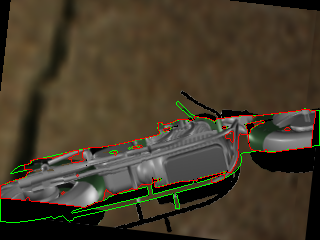}}
		\\ \vspace{0.05cm}
		$\bI(\bxi^{\star}_{\text{final}})$ & 
		\parbox[c]{0.20\textwidth}{\includegraphics[width=0.22\textwidth]{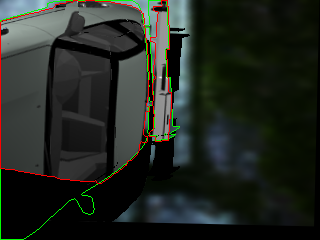}}&
		\parbox[c]{0.20\textwidth}{\includegraphics[width=0.22\textwidth]{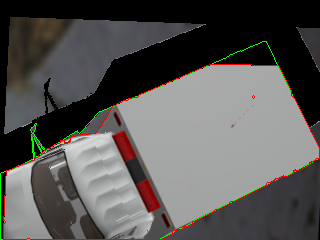}}&
		\parbox[c]{0.20\textwidth}{\includegraphics[width=0.22\textwidth]{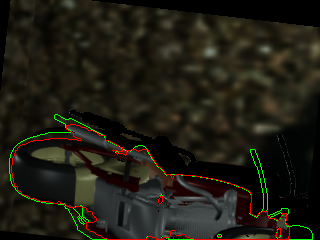}}&
		\parbox[c]{0.20\textwidth}{\includegraphics[width=0.22\textwidth]{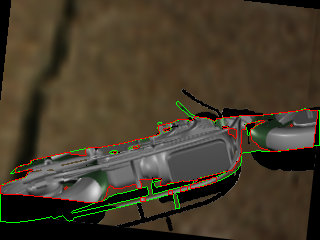}}
		\\ \vspace{0.05cm}
		$\bI(\bxi^{\text{GT}})$ & 
		\parbox[c]{0.20\textwidth}{\includegraphics[width=0.22\textwidth]{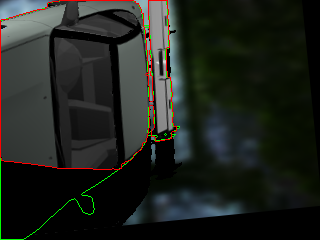}}&
		\parbox[c]{0.20\textwidth}{\includegraphics[width=0.22\textwidth]{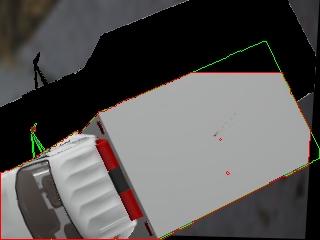}}&
		\parbox[c]{0.20\textwidth}{\includegraphics[width=0.22\textwidth]{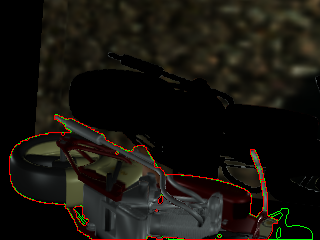}}&
		\parbox[c]{0.20\textwidth}{\includegraphics[width=0.22\textwidth]{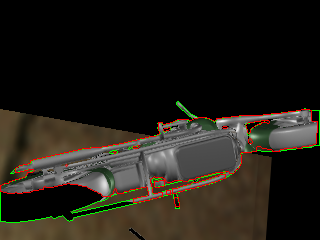}}
		\\ 	
	\end{tabularx}
	\caption{\textbf{Qualitative Results on MovingObjects3D.} Visualization of the warped image $\bI(\bxi)$ using the ground truth object motion $\bxi^{\text{GT}}$ (last row) and the object motion $\bxi^{\star}$ estimated using our method at each pyramid ($\bxi^{\star}_{1}$, $\bxi^{\star}_{2}$, $\bxi^{\star}_{3}$, $\bxi^{\star}_{\text{final}}$) on the MovingObjects3D \emph{validation} and \emph{test} set. In $\bI$, we plot the instance boundary in \textcolor{red}{red} and that of $\bT$ in \textcolor{darkgreen}{green} as comparison. Note the difficulty of the task (truncation, independent background object) and the high quality of our alignments. Black regions in the warped image are due to truncation or occlusion.}
	\label{fig:iterative_vis2}
\end{figure*}

\begin{figure*}[t]
	\centering 
	\begin{tabularx}{\textwidth}{ccccc}
		\vspace{0.05cm}
		\vspace{0.05cm}
		$\bT$ & 
		\parbox[c]{0.18\textwidth}{\includegraphics[width=0.20\textwidth]{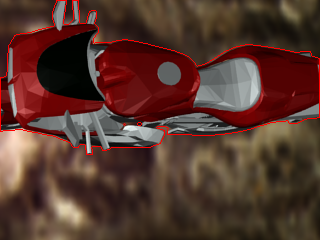}}&
		\parbox[c]{0.18\textwidth}{\includegraphics[width=0.20\textwidth]{results/iterations_visualization/plane192_22to24/T_vis.png}}&
		\parbox[c]{0.18\textwidth}{\includegraphics[width=0.20\textwidth]{results/iterations_visualization/plane192_26to28/T_vis.png}}&
		\parbox[c]{0.18\textwidth}{\includegraphics[width=0.2\textwidth]{results/iterations_visualization/car192_28to32/T_vis.png}}
		\\ \vspace{0.05cm}
		$\bI$ & 
		\parbox[c]{0.18\textwidth}{\includegraphics[width=0.20\textwidth]{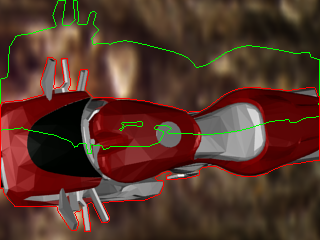}}&
		\parbox[c]{0.18\textwidth}{\includegraphics[width=0.20\textwidth]{results/iterations_visualization/plane192_22to24/I_vis.png}}&
		\parbox[c]{0.18\textwidth}{\includegraphics[width=0.20\textwidth]{results/iterations_visualization/plane192_26to28/I_vis.png}}&
		\parbox[c]{0.18\textwidth}{\includegraphics[width=0.20\textwidth]{results/iterations_visualization/car192_28to32/I_vis.png}}
		\\ \vspace{0.05cm}
		\begin{tabular}{c} $\bI(\bxi^{\star})$ \\ DeepLK \\ 6DoF \cite{Wang18icra} \end{tabular} & 
		\parbox[c]{0.18\textwidth}{\includegraphics[width=0.20\textwidth]{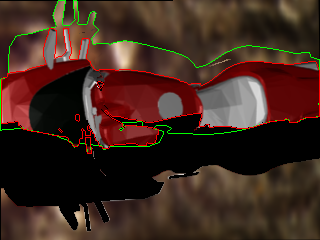}}&
		\parbox[c]{0.18\textwidth}{\includegraphics[width=0.20\textwidth]{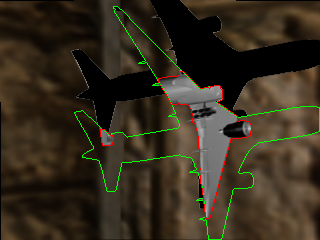}}&
		\parbox[c]{0.18\textwidth}{\includegraphics[width=0.20\textwidth]{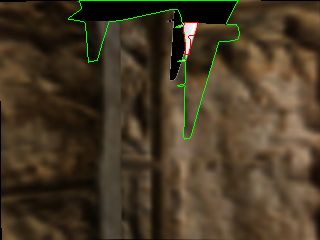}}&
		\parbox[c]{0.18\textwidth}{\includegraphics[width=0.20\textwidth]{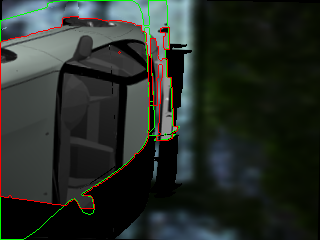}}
		\\ \vspace{0.05cm}
		\begin{tabular}{c} $\bI(\bxi^{\star})$ \\ Ours: (A) \end{tabular} & 
		\parbox[c]{0.18\textwidth}{\includegraphics[width=0.20\textwidth]{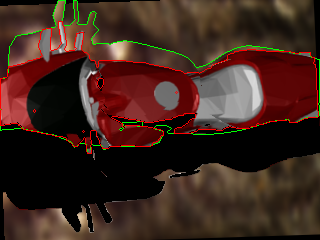}}&
		\parbox[c]{0.18\textwidth}{\includegraphics[width=0.20\textwidth]{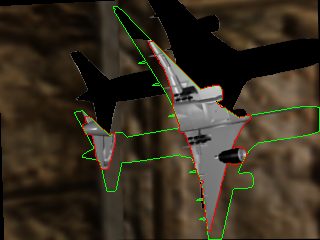}}&
		\parbox[c]{0.18\textwidth}{\includegraphics[width=0.20\textwidth]{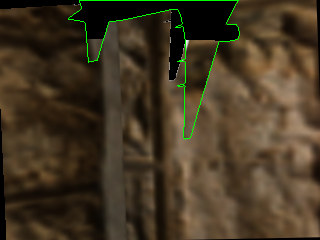}}&
		\parbox[c]{0.18\textwidth}{\includegraphics[width=0.20\textwidth]{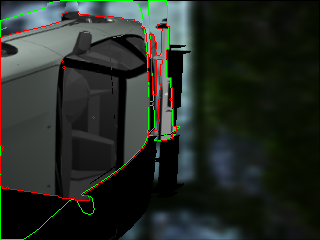}}
		\\ \vspace{0.05cm}
		\begin{tabular}{c} $\bI(\bxi^{\star})$ \\ Ours: \\(A)+(B) \end{tabular} &		\parbox[c]{0.18\textwidth}{\includegraphics[width=0.20\textwidth]{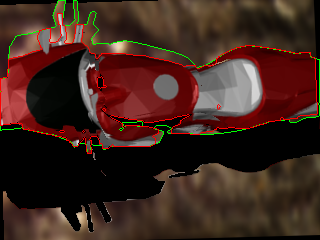}}&
		\parbox[c]{0.18\textwidth}{\includegraphics[width=0.20\textwidth]{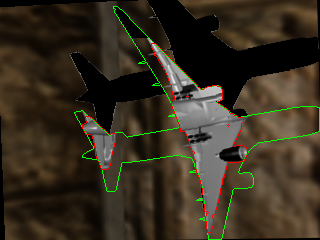}}&
		\parbox[c]{0.18\textwidth}{\includegraphics[width=0.20\textwidth]{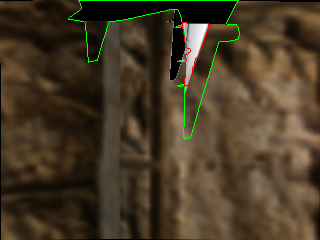}}&
		\parbox[c]{0.18\textwidth}{\includegraphics[width=0.20\textwidth]{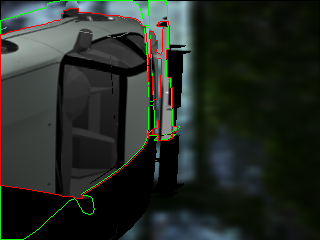}}
		\\ \vspace{0.05cm}
		\begin{tabular}{c} $\bI(\bxi^{\star})$ \\ Ours: (A)+\\(B)+(C) \end{tabular} &		\parbox[c]{0.18\textwidth}{\includegraphics[width=0.20\textwidth]{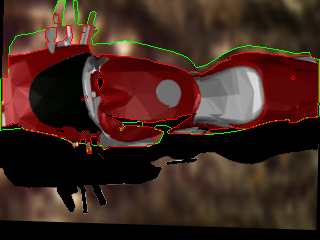}}&
		\parbox[c]{0.18\textwidth}{\includegraphics[width=0.20\textwidth]{results/iterations_visualization/plane192_22to24/ours_vis.png}}&
		\parbox[c]{0.18\textwidth}{\includegraphics[width=0.20\textwidth]{results/iterations_visualization/plane192_26to28/ours_vis.png}}&
		\parbox[c]{0.18\textwidth}{\includegraphics[width=0.20\textwidth]{results/iterations_visualization/car192_28to32/ours_vis.png}}
		\\ \vspace{0.05cm}
		$\bI(\bxi^{\text{GT}})$ & 
		\parbox[c]{0.18\textwidth}{\includegraphics[width=0.20\textwidth]{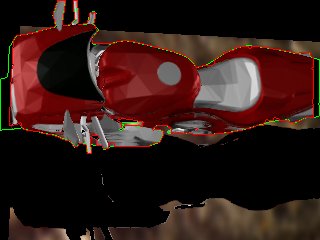}}&
		\parbox[c]{0.18\textwidth}{\includegraphics[width=0.20\textwidth]{results/iterations_visualization/plane192_22to24/GT_vis.png}}&
		\parbox[c]{0.18\textwidth}{\includegraphics[width=0.20\textwidth]{results/iterations_visualization/plane192_26to28/GT_vis.png}}&
		\parbox[c]{0.18\textwidth}{\includegraphics[width=0.20\textwidth]{results/iterations_visualization/car192_28to32/GT_vis.png}}
		\\ 	
	\end{tabularx}
	\caption{\textbf{Qualitative Comparisons of Our Method to Baselines on MovingObjects3D.} We compared the object motion $\bxi^{\star}$ estimated using our proposed modules (A)+(B)+(C) $\bxi^{\star}$ (row 6) to the optimal poses output from DeepLK 6DoF (row 3), ours (A) (row 4) and ours (A)+(B) (row 5) on the MovingObjects3D \emph{validation} and \emph{test} set. We visualize the warped image $\bI(\bxi)$ using the ground truth object motion $\bxi^{\text{GT}}$ (last row). In $\bI$, we plot the instance boundary in \textcolor{red}{red} and that of $\bT$ in \textcolor{darkgreen}{green} for qualitative comparison of the two shapes in 2D. }
	\label{fig:method_comparison1}
\end{figure*}

We now demonstrate additional qualitative results on 3D rigid transformation estimation on MovingObjects3D. Note the difficulty of the task (truncation, independent background object) and the high quality of our alignments. 

\boldparagraph{Visualizations of Iterative Estimation} \figref{fig:iterative_vis1} and \figref{fig:iterative_vis2} show a qualitative visualization of our method at different iterations. We warp the image $\bI$ using the iterative estimated poses at the four coarse-to-fine pyramid scales ($\bI(\bxi^{\star}_{1}) \rightarrow \bI(\bxi^{\star}_{2}) \rightarrow \bI(\bxi^{\star}_{3}) \rightarrow \bI(\bxi^{\star}_{\text{final}})$ ). $\bI(\bxi^{\star}_{\text{final}})$ is the final output of our method. Our results demonstrate that the proposed method is able to iteratively refine the estimation towards the global optimal solution despite the challenging scenario.

\boldparagraph{Comparison to Baselines} \figref{fig:method_comparison1} shows a comparison of our full method to DeepLK 6DoF \cite{Wang18icra} and ablated models using only parts of the proposed modules. Our results demonstrate that the combination of all modules yields high-quality registrations.




\end{document}